\documentclass[lettersize,journal]{IEEEtran}
\usepackage{amsmath,amsfonts}
\usepackage{algorithmic}
\usepackage{algorithm}
\usepackage{array}
\usepackage[caption=false,font=normalsize,labelfont=sf,textfont=sf]{subfig}
\usepackage{textcomp}
\usepackage{booktabs} 
\usepackage{multirow}
\usepackage{stfloats}
\usepackage{url}
\usepackage{verbatim}
\usepackage{graphicx}
\usepackage{cite}
\usepackage{bm}
\usepackage{makecell}
\usepackage{arydshln}
\usepackage[pagebackref=true,breaklinks=true,colorlinks,bookmarks=false,citecolor=blue,linkcolor=blue]{hyperref}

\begin{document}

\title{Deep Unrestricted Document Image Rectification}

\author{Hao Feng$^{\dagger}$,
 Shaokai Liu$^{\dagger}$,
 Jiajun Deng,
 Wengang Zhou*,~\IEEEmembership{Senior Member,~IEEE},
 and~Houqiang Li*,~\IEEEmembership{Fellow,~IEEE}
 
\IEEEcompsocitemizethanks{
\IEEEcompsocthanksitem This work was supported in part by the National Natural Science Foundation of China under Contract U20A20183 and 62021001. It was also supported by the GPU cluster built by MCC Lab of Information Science and Technology Institution and the Supercomputing Center of the USTC.
\IEEEcompsocthanksitem Hao Feng, Shaokai Liu, Wengang Zhou, and Houqiang Li are with the CAS Key Laboratory of Technology in Geo-spatial Information Processing and Application System, Department of Electronic Engineering and Information Science, University of Science and Technology of China, Hefei, 230027, China.
Hao Feng is also with Zhangjiang Laboratory, Shanghai, China.
E-mail: \{haof, liushaokai\}@mail.ustc.edu.cn; \{zhwg, lihq\}@ustc.edu.cn
\IEEEcompsocthanksitem Jiajun Deng is with The University of Adelaide, Australian Institute for Machine Learning. E-mail: jiajun.deng@adelaide.edu.au
\IEEEcompsocthanksitem  $^{\dagger}$The first two authors contribute equally to this work.
\IEEEcompsocthanksitem  *Corresponding authors: Wengang Zhou and Houqiang Li.
}}



\maketitle

\begin{abstract}
In recent years, tremendous efforts have been made on document image rectification, 
but existing advanced algorithms are limited to processing restricted document images, 
\emph{i.e.}, the input images must incorporate a complete document. 
Once the captured image merely involves a local text region,
its rectification quality is degraded and unsatisfactory. 
Our previously proposed DocTr, a transformer-assisted network for document image rectification, also suffers from this limitation.
In this work, we present DocTr++, 
a novel unified framework for document image rectification,
without any restrictions on the input distorted images.
Our major technical improvements can be concluded in three aspects.
Firstly,
we upgrade the original architecture by adopting a hierarchical encoder-decoder structure for multi-scale representation extraction and parsing.
Secondly,
we reformulate the pixel-wise mapping relationship between the unrestricted distorted document images and the distortion-free counterparts.
The obtained data is used to train our DocTr++ for unrestricted document image rectification.
Thirdly, we contribute a real-world test set and metrics applicable for evaluating the rectification quality.
To our best knowledge, this is the first learning-based method for the rectification of unrestricted document images.
Extensive experiments are conducted, 
and the results demonstrate the effectiveness and superiority of our method.
We hope our DocTr++ will serve as a strong baseline for generic document image rectification,
prompting the further advancement and application of learning-based algorithms.
The source code and the proposed dataset are publicly available at \url{https://github.com/fh2019ustc/DocTr-Plus}.
\end{abstract}


\section{Introduction}
\IEEEPARstart{N}owadays, the widespread use of smartphones has led to a growing trend of directly using them for digitizing document files. 
Compared with traditional flatbed scanners,
smartphones provide a more flexible, portable, and straightforward alternative for document image digitalization.
However, 
those captured document images are inevitably distorted,
due to some uncontrollable factors, including physical deformations of documents, illumination conditions, and camera angles. 
Such distortions block their digital storage and are likely to negatively impact downstream applications, such as automatic text recognition~\cite{ciardiello1988experimental,wang2011end,lat2018enhancing,peng2022recognition}, analysis~\cite{yuan2019interpreting,zhang2022multimodal,kim2022ocr}, retrieval~\cite{salton1991developments,yang2008harmonizing,liu2023end}, and question answering~\cite{mathew2021docvqa,nie2012beyond,feng2023unidoc}.
Over the past few decades,
document image rectification has been actively researched. 

\begin{figure}[t]
  \centering
  \includegraphics[width=0.94\linewidth]{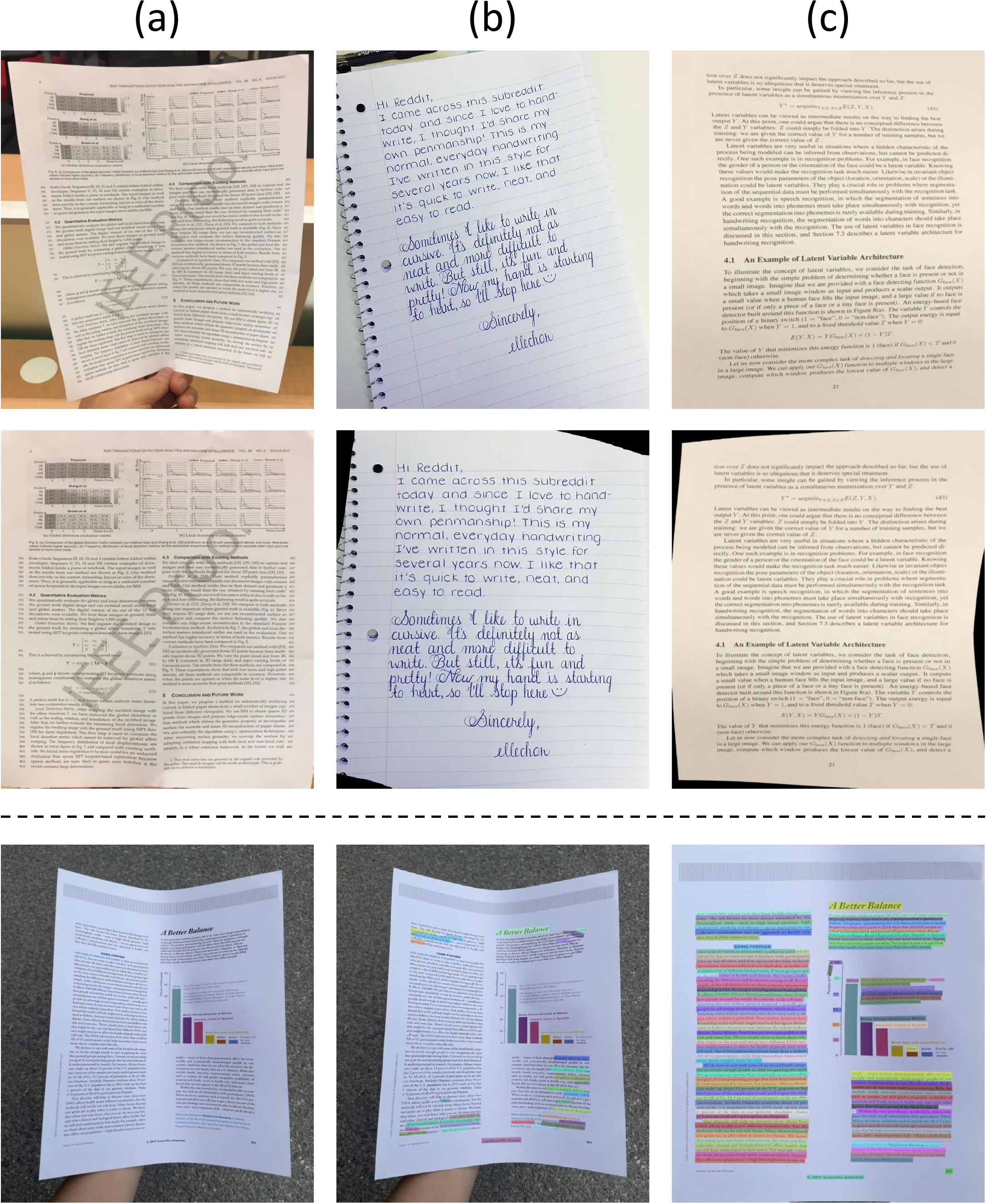}
  \caption{Top row: three types of commonly distorted document images based on the presence of document boundaries: (a) w/ complete boundaries, (b) w/ partial boundaries, and (c) w/o any boundaries. Middle row: the rectified results of our method. Bottom row: the distorted image, the original detected texts, and the rectified one (highlighted), based on DBNet~\cite{liao2020real,CnSTD2023}.}
  \label{fig:doc_class}
\end{figure}

In the literature, the conventional solutions are based on 3D reconstruction techniques.
Generally, they construct a 3D representation of the captured distorted document and then flatten it into a distortion-free planar.
To perform 3D reconstruction, some approaches resort to the auxiliary hardwares, such as structured-lighting system~\cite{brown2004image,937649,6909892} or laser range scanner~\cite{4407722}.
Other methods~\cite{4916075,yamashita2004shape,tsoi2007multi,tsoi2004geometric,Shaodi} capture multiview images of a document as an alternative.
However, the extra hardwares and the demand for multiview images inevitably limit their usage in real applications.
To overcome these limitations,
some other methods assume a parametric model on the deformed surface of documents 
and optimize the model by extracting specific attributes, such as
shading~\cite{1561180,wada1997shape}, boundaries~\cite{6628653}, textlines~\cite{1227630,958227,wu2002document,meng2011metric,luo2022geometric}, or texture flow~\cite{liang2008geometric}.
Nevertheless, the oversimplified parametric models 
usually lead to limited performance, and the optimization process introduces non-negligible computational cost.

Recently, deep learning based solutions~\cite{das2020intrinsic,9010747,das2021end,feng2021doctr,jiang2022revisiting,liu2020geometric,8578592,xie2021document,xie2020dewarping,xue2022fourier,feng2021docscanner,feng2022geometric,markovitz2020can,ma2022learning,zhang2022marior,li2019document} have been shown as an alternative to the traditional methods, with promising performance and efficiency.
By inferring a pixel-wise displacement field~\cite{teed2020raft} from the distorted image to the distortion-free one with a deep network, a distorted document image can be unwarped with the predicted flow field at a high speed.
However, most existing algorithms~\cite{das2020intrinsic,9010747,das2021end,feng2021doctr,jiang2022revisiting,liu2020geometric,8578592,xie2021document,xie2020dewarping,xue2022fourier,feng2021docscanner,feng2022geometric,markovitz2020can,zhang2022marior,ma2022learning} are dedicated to processing restricted document images, \emph{i.e.}, the input images must contain complete document boundaries (see Fig.~\ref{fig:doc_class}(a)). 
Once the distorted images do not include complete document boundaries (see Fig.~\ref{fig:doc_class}(b, c)), the rectification quality is degraded.
To overcome this issue, DocProj~\cite{li2019document} divides the input images into multiple patches and infers per-patch displacement field~\cite{teed2020raft}.
However, the stitching of the per-patch displacement field involves heavy computational cost~\cite{freedman2005interactive} and this method is limited to rectifying the document images without any background regions (see Fig.~\ref{fig:doc_class}(c)).
For document images with some background regions, we need to manually cut out the document region first before the rectification.
Therefore, a robust algorithm capable of rectifying various distorted document images in the wild is being sought.

As shown in Fig.~\ref{fig:doc_class},
based on the existence of a complete document, a distorted document image can be divided into three categories, \emph{i.e.}, 
(a) with complete boundaries, (b) with partial boundaries, and (c) with no boundaries.
To adaptively rectify these cases with a unified model, in this study, we propose DocTr++, a novel deep network for unrestricted document image rectification.
Our DocTr++ extends our previously proposed DocTr~\cite{feng2021doctr} that focuses on rectifying images containing complete documents,
and makes improvements in the following three aspects.
Firstly, 
we upgrade the architecture and adopt a hierarchical encoder-decoder structure for multi-scale representation learning and parsing, achieving improved distortion rectification.
Secondly, to facilitate our motivation for rectifying any distorted document images,
we reformulate the pixel-wise mapping relationship between the unrestricted distorted document images and their distortion-free counterparts.
Thirdly,
we contribute a real-world benchmark and metrics to evaluate the rectification quality in this scenario.
To our best knowledge, this is the first learning-based method and dataset for rectifying unrestricted document images.

Extensive experiments are conducted on the challenging DocUNet Benchmark dataset~\cite{8578592} and on our proposed benchmark dataset.
The quantitative and qualitative results showcase the effectiveness and superiority of our method over existing solutions. 
We hope that our DocTr++ will serve as a strong baseline for unrestricted document image rectification,
encouraging further advancement and application of learning-based algorithms for document image rectification.

In summary, we make three-fold contributions as follows:
\begin{itemize}
    \item
    We make the first attempt to unrestricted document image rectification, and propose a novel solution, \emph{i.e.,} DocTr++, adopting a hierarchical encoder-decoder structure for effective representation encoding and parsing.
    \item
    We contribute a new document image dataset and applicable metrics to facilitate the training and evaluation for unrestricted document image rectification.
    \item
    We conduct extensive experiments to validate the merits of our method and
    set several new state-of-the-art records on the prevalent and proposed benchmarks.
\end{itemize}

\section{Related Work}
In this section, we first systematically classify prior work on document image rectification into two categories: (a) 3D reconstruction-based methods and (b) deep learning-based approaches, discussed next.

\subsection{Rectification Based on 3D Reconstruction} 
Traditional solutions for document image rectification rely on 3D reconstruction techniques. These methods~\cite{brown2004image,937649,6909892,4407722} commonly first reconstruct the 3D shape of a deformed document page and then flatten it to a planar shape.
Typically, Brown and Seales~\cite{937649} first employed a structured light 3D acquisition system for 3D model reconstruction.
Zhang~et al.~\cite{4407722} adopted a laser range scanner instead and performed restoration based on a physical modeling technique. 
Meng~et al.~\cite{6909892} utilized two structured beams to illuminate upon the deformed document page and obtained its developable surface for rectification.
While these methods have shown effectiveness in real-world scenarios, they require additional hardware to scan the deformed documents, making them less practical for personal use.
	
Some methods work from multiview images of a document for 3D reconstruction.
Typically, Tsoi~et al.~\cite{tsoi2007multi} convert the multiview images to a canonical coordinate frame based on the document boundary interpolation~\cite{tsoi2004geometric}.
Koo~et al.~\cite{4916075} match the corresponding points in two views by SIFT~\cite{lowe2004distinctive} to estimate the document surface model.
Based on the ridge-aware 3D reconstruction techniques, You~et al.~\cite{Shaodi} propose a general developable surface model that can represent a wide variety of paper deformations.
For such above works, the complicated image acquisition steps still limit their further applications.

To address the above issues, other methods reconstruct the 3D model from a single-view image.
They first assume a parametric model on the document surface and then fit the model by extracting the specific representations, 
including
shading~\cite{1561180,wada1997shape}, boundaries~\cite{6628653}, textlines~\cite{1227630,958227,wu2002document,meng2011metric,luo2022geometric}, or texture flow~\cite{liang2008geometric}.
Typically,
using a proximal light source, Wada et al.~\cite{wada1997shape} recover the 3D shape of an unfolded book surface based on the shading distribution. 
He et al.~\cite{6628653} extract a book boundary model for the reconstruction of 3D book surface. 
Textline-based method~\cite{1227630,958227,wu2002document,meng2011metric} segment the textlines in distorted document pages to estimate the parameters of the surface model.
Besides, Meng et al.~\cite{meng2018exploiting} recover the 3D page shape by exploiting the intrinsic vector fields of the image.
However, these methods still have some limitations. On the one hand, the fixed parametric models are not usually applicable to the complicated real geometric distortions of deformed documents. On the other hand, the optimization process
introduces non-negligible computational cost.

\subsection{Rectification Based on Deep Learning}
Deep learning has been introduced recently and shown as a promising alternative to the conventional works.
By predicting a pixel-wise displacement field~\cite{teed2020raft} with a deep network, 
a distorted document image can be rectified by resampling the pixels from the input distorted image to the rectified one.

As a pioneering work, DocUNet~\cite{8578592} regresses a pixel-wise displacement field with a stacked UNet~\cite{ronneberger2015u} to unwarp the input distorted document image.
DocProj~\cite{li2019document} divides the input distorted document image into multiple patches, estimates the per-patch displacement field, and finally stitches them for full image rectification.
Using a fully convolutional network~\cite{long2015fully}, Xie~et al.~\cite{xie2020dewarping} supplement an explicit regularization to control the smoothness of the displacement field.
Inspired by the traditional methods based on 3D reconstruction, 
Amir~et al.~\cite{markovitz2020can} and Das~et al.~\cite{9010747} model the 3D shape of a deformed document in the network.
DocTr~\cite{feng2021doctr} introduces transformer~\cite{Vaswani2017AttentionIA} from natural language processing tasks to improve the representation learning of document images. 

Different from the pixel-wise flow regression, DDCP~\cite{xie2021document} only estimates several control points and reference points to perform rectification,
based on the TPS interpolation algorithm~\cite{meijering2002chronology}.
PWUNet~\cite{das2021end} considers the different distortion degrees of local patches and estimates local deformation fields for improved global unwarping.
DocGeoNet~\cite{feng2022geometric} proposes to encode representations of rectification cues to improve rectification, by detecting textlines in the network.
RDGR~\cite{jiang2022revisiting} first detect textlines and document boundaries, and then obtain the rectified image by solving an optimization problem based on the proposed grid regularization.
FDRNet~\cite{xue2022fourier} concentrates on high-frequency components in the Fourier space to capture structural information of deformed documents for improved rectification.
Marior~\cite{zhang2022marior} considers the document images with large marginal regions and gradually rectifies them to a robust state.
DocScanner~\cite{feng2021docscanner} proposes to rectify the distortion in a progressive and iterative manner with a lightweight network.
PaperEdge~\cite{ma2022learning} boosts document unwarping performance by introducing real-world document images in the training.
CGU-Net~\cite{verhoeven2023neural} introduces a novel grid-based method for document unwarping and contributes a high-quality synthetic dataset.

\begin{figure*}[t]
  \centering
  \includegraphics[width=1\linewidth]{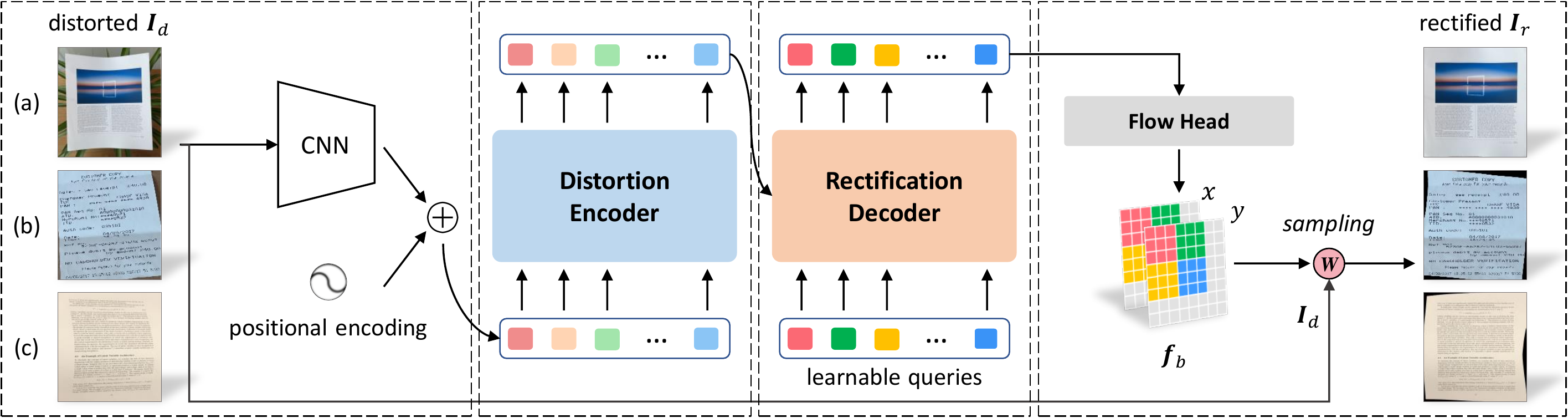}
  \caption{An overview of our DocTr++ for unrestricted document image rectification. Given an arbitrary distorted document image $\bm{I}_d$, we extract its features through a CNN backbone and a distortion encoder architecture. 
  Then, the rectification decoder takes a fixed number of learned queries as input that attend to the encoder's output. 
  These embeddings are parallelly transformed into per-patch warping flows $\bm{f}_b$ pointing to $\bm{I}_d$.
  Finally, we use the predicted $\bm{f}_b$ to warp $\bm{I}_d$ and obtain the rectified image $\bm{I}_r$ through the bilinear sampling-based warping operation ``W".
  }
  \label{fig:overview}
\end{figure*}

Although the above learning-based works are reported with state-of-the-art performance and running efficiency,
they are still limited to processing document images with a controlled appearance.
Among them, methods~\cite{9010747,das2021end,feng2021doctr,jiang2022revisiting,liu2020geometric,8578592,xie2021document,xie2020dewarping,xue2022fourier,feng2021docscanner,feng2022geometric,markovitz2020can,li2019document,zhang2022marior,ma2022learning} can only process images with complete document and the captured documents are surrounded by a limited size background area.
Besides, Marior~\cite{zhang2022marior} innovatively proposes to rectify the images with larger background areas, while DocProj~\cite{li2019document} mainly considers the samples with no background regions.
In contrast, in this work, we aim to achieve an adaptive rectification of various document images in the wild, regardless of whether or not they incorporate complete document boundaries.
We hope that our method will provide a strong baseline and attract new researchers to the field, prompting the real applications of document image rectification algorithms.

\section{Approach}
Fundamentally, the task of document image rectification can be considered as a monocular optical flow estimation task. In standard optical flow estimation~\cite{dosovitskiy2015flownet}, the input consists of two consecutive frames, and the output describes the pixel-level correspondences between these frames. In the document image rectification task, the input is a distorted document image and the network aims to predict the pixel-level correspondences from this image to a distortion-free one. Based on the output correspondences, pixel displacements are performed, resulting in a rectified distortion-free document image.

In this study, we first expanded all possible types of input distorted document images, including (a) with complete document boundaries, (b) with partial document boundaries, and (c) without any document boundaries.
Furthermore, in contrast to the original DocTr~\cite{feng2021doctr}, which operates on feature maps scaled down to 1/8 of the input image size, we develop a multi-scale feature fusion approach to reduce computational costs while ensuring rectification accuracy. In our network design, we build upon the classical transformer~\cite{Vaswani2017AttentionIA}, driven by motivations from the following two perspectives. On one hand, the self-attention mechanism within the encoder of transformer~\cite{Vaswani2017AttentionIA} is adept at extracting structural features of the deformed paper sheet, given the long-range dependencies of elements like textlines across document images. On the other hand, the cross-attention mechanism within the decoder of transformer~\cite{Vaswani2017AttentionIA} facilitates effective querying and reorganization of features extracted by the encoder, serving the purpose of rectified image prediction.

Fig.~\ref{fig:overview} shows an overview of our DocTr++. 
Given a distorted document image $\bm{I}_d$,
we first utilize a conventional CNN backbone~\cite{he2016deep} to extract its 2D representation and feed it to a distortion encoder~\cite{Vaswani2017AttentionIA}.
Then, a rectification decoder~\cite{Vaswani2017AttentionIA} takes as input a fixed number of learned embeddings that attend to the output of the encoder,
and decodes per-patch warping flow $\bm{f}_b$ pointing at pixels in distorted image $\bm{I}_d$. With the predicted $\bm{f}_b$, we sample the pixels from image $\bm{I}_d$ to reconstruct the rectified image $\bm{I}_r$.
Different from previous methods, for our DocTr++ the input document images $\bm{I}_d$ cover various situations captured in daily life, including (a) with complete document boundaries, (b) with partial document boundaries, and (c) without any document boundaries. 
In the following, we separately discuss the four main components of our DocTr++, including (a) backbone, (b) distortion encoder, (c) rectification decoder, and (d) flow head.

\subsection{Backbone}
Starting from an arbitrary distorted document image $\bm{I}_d \in\mathbb{R}^{H\times W\times3}$, where $H$ and $W$ denote the height and width of this \emph{RGB} image, we first extract its features with a convolutional module. This convolutional module consists of 6 residual blocks~\cite{he2016deep}. It downsamples the feature maps at $\frac{1}{2}$ resolution every two residual blocks and finally produce feature map $\bm{E}_{c} \in\mathbb{R}^{\frac{H}{8}\times\frac{W}{8}\times C_{b}}$, where we set $C_{b} = 256$.
                                                               
\subsection{Distortion Encoder}
To rectify a distorted document image, it is essential to capture its structural information.
However, in a document image, such information usually exists in components distributed among non-local regions, such as curved textlines and textures with gradually changing light appearance.
To encode the structural information for an improved rectification,
we propose to introduce the self-attention mechanism from transformer~\cite{Vaswani2017AttentionIA} and construct a hierarchical distortion encoder.

\begin{figure}[t]
  \centering
  \includegraphics[width=1\linewidth]{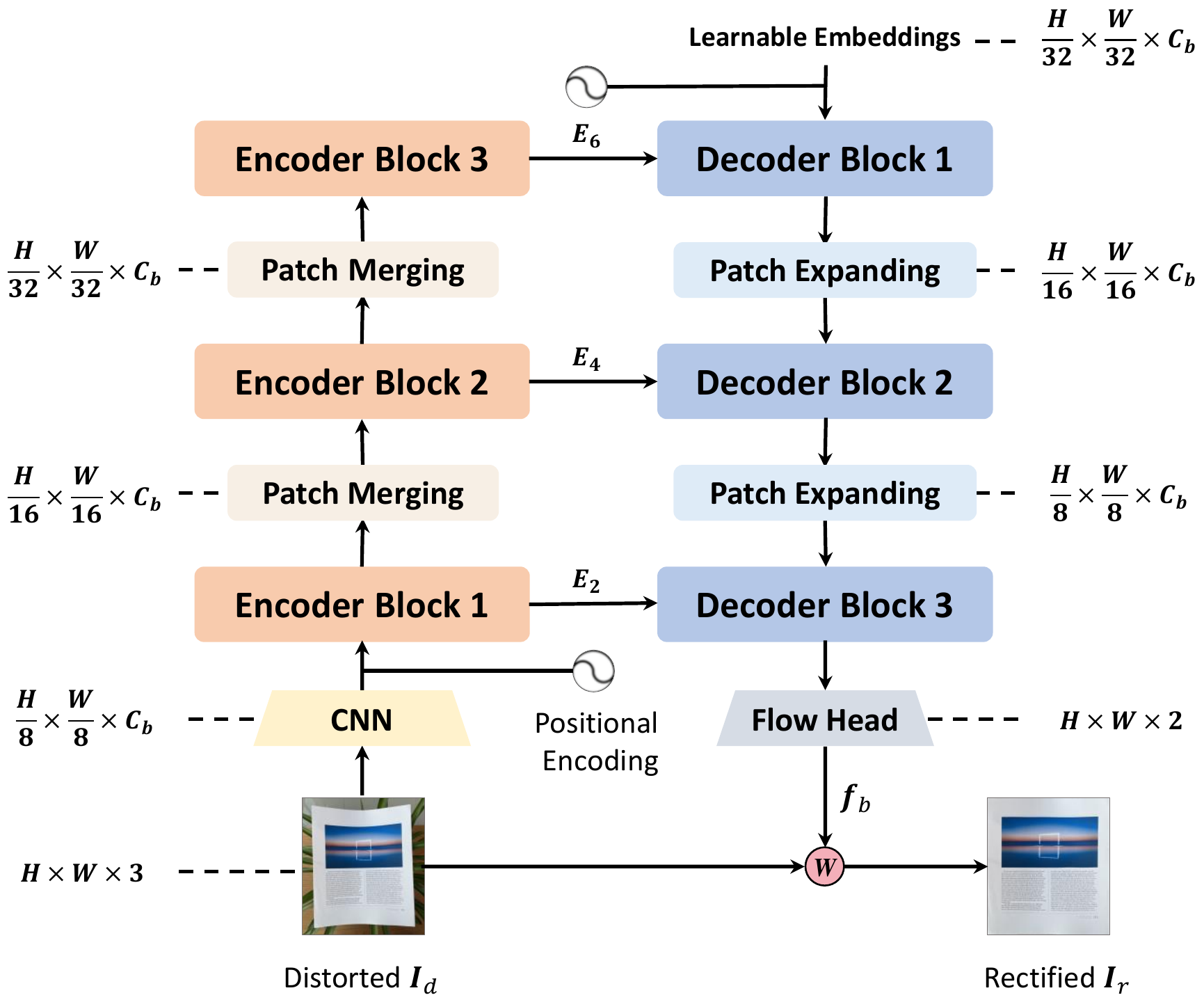}
  \caption{Detailed architecture of the proposed DocTr++. It takes a hierarchical architecture that incorporates a distortion encoder for multi-level feature extraction and a rectification decoder for warping flow prediction.
  }
  \label{fig:tr}
\end{figure}

Specifically, 
as shown in Fig.~\ref{fig:tr},
the distortion encoder consists of three blocks and each block contains two standard transformer encoder layers~\cite{Vaswani2017AttentionIA}.
To generate both high-resolution fine features and low-resolution semantically stronger features,
we downsample the feature maps at $\frac{1}{2}$ resolution with a convolutional layer ($stride=2$) after the first and second blocks.
Since the transformer architecture~\cite{Vaswani2017AttentionIA} is permutation-invariant,
we supplement the image features with the fixed 2D position embedding $\bm{P}_e$~\cite{bello2019attention} before the transformer encoder layer.
Each layer contains a multi-head self-attention module and a feed-forward network.
The output representation in each layer can be calculated as follows,
\begin{equation}
\begin{aligned}
    \bm{E}_0 &= \bm{E}_{c}+\bm{P}_{e}, \\
    \bm{E}^{'}_i &= LN(MA(\bm{E}_{i-1}, \bm{E}_{i-1}) + \bm{E}_{i-1}), i=\{1,...,6\}, \\
    \bm{E}_i &= LN(FFN(\bm{E}^{'}_i) + \bm{E}^{'}_i), i=\{1,...,6\},
\end{aligned}
\label{equ:transenc}
\end{equation}
where $MA(\cdot)$, $FFN(\cdot)$, and $LN(\cdot)$ are the multi-head attention, feed-forward network, and layer normalization, respectively,
$\bm{E}_i$ denotes the output feature of the $i^{th}$ encoder layer.

\subsection{Rectification Decoder}
As shown in Fig.~\ref{fig:tr}, the rectification decoder takes as input the hierarchical feature maps $\{\bm{E}_2, \bm{E}_4, \bm{E}_6\}$ from the distortion encoder as well as a
small fixed number of learned embeddings $\bm{D}_l \in \mathbb{R}^{(\frac{H}{8}\times\frac{W}{8})\times C_{b}}$.
It outputs the features for predicting the warping flow $\bm{f}_b \in \mathbb{R}^{H\times W\times 2}$ used to rectify the input distorted image $\bm{I}_d$.
We term the learned embeddings as the rectification queries and each query is responsible for the rectification of a corresponding region in the input distorted image.

The rectification decoder consists of three blocks and each block includes two standard transformer decoder layers~\cite{Vaswani2017AttentionIA}.
For the first block, we take as input the learnable embedding $\bm{D}_l \in \mathbb{R}^{\frac{H}{32}\times \frac{W}{32}\times C_b}$ to attend to the feature map $\bm{E}_6 \in \mathbb{R}^{\frac{H}{32}\times \frac{W}{32}\times C_b}$.
For the other two blocks, we take the currently decoded embeddings as input to attend to the feature map $\bm{E}_4 \in \mathbb{R}^{\frac{H}{16}\times \frac{W}{16}\times C_b}$ and $\bm{E}_2 \in \mathbb{R}^{\frac{H}{8}\times \frac{W}{8}\times C_b}$, respectively.
Note that the decoded embeddings of the first and second blocks are upsampled based on the bilinear interpolation when fed into the next block.
Each decoder layer consists of a multi-head self-attention module, an encoder-decoder multi-head attention module, and a feed-forward network. The formula in each layer is given as follows,
\begin{equation}
\begin{aligned}
    \bm{D}_0&=\bm{D}_{l}+\bm{P}_{d},\\
    \bm{D}_i'&=LN({MA}(\bm{D}_{i-1}, \bm{D}_{i-1})+\bm{D}_{i-1}), i=\{1,...,6\}, \\
    \bm{D}_i''&= LN({MA}(\bm{D}_i', \bm{E}_k)+\bm{D}_i'), i=\{1,...,6\}, k=\{2,4,6\},\\
    \bm{D}_i&=LN({FFN}(\bm{D}_i'')+\bm{D}_i''),
\end{aligned}
\end{equation}
where $\bm{D}_i$ indicates the output of the $i^{th}$ decoder layer, $\bm{P}_{d}$ represents the positional embeddings~\cite{bello2019attention}, and $\bm{E}_k$ denotes the attended features from the corresponding encoder block.

\subsection{Flow Head}
Given the decoded features $\bm{D}_6 \in \mathbb{R}^{\frac{H}{8}\times \frac{W}{8}\times C_b}$,
we introduce a learnable upsampling module~\cite{feng2021doctr} to predict the warping flow $\bm{f}_b \in \mathbb{R}^{H \times W\times 2}$ for rectifying the distorted image $\bm{I}_d$.

Specifically, we first apply two convolutional layers to the feature map $\bm{D}_6 \in \mathbb{R}^{\frac{H}{8} \times \frac{W}{8} \times C_b}$ and produce the coarse warping flow 
$\bm{f}_m \in \mathbb{R}^{\frac{H}{8}\times \frac{W}{8}\times 2}$.
Then, we exploit another two convolutional layers to process $\bm{D}_6 \in \mathbb{R}^{\frac{H}{8} \times \frac{W}{8} \times C_b}$, and reshape the output feature map to shape $\frac{H}{8} \times \frac{W}{8} \times 8\times8\times9$.
Next, we perform softmax on the last dimension of it and get the weight matrix.
Using the obtained weight matrix, we take a weighted combination over the $3\times3$ neighborhood of each pixel in $\bm{f}_m$.
Finally, the obtained $\frac{H}{8} \times \frac{W}{8} \times 8\times8 \times 2$ map is permuted and reshaped to the full resolution warping flow map $\bm{f}_b \in \mathbb{R}^{H \times W \times 2}$.
With the predicted $\bm{f}_b$, the rectified document image $\bm{I}_r\in\mathbb{R}^{H\times W\times 3}$ can be obtained by a warping operation based on the bilinear interpolation as follows,
\begin{equation}\label{equ:task}
	\bm{I}_r(u_0,v_0) = \bm{I}_d(\bm{f}_b^{u}(u_0,v_0), \bm{f}_b^{v}(u_0,v_0)),
\end{equation}
where $\bm{f}_b^{u} \in \mathbb{R}^{H \times W \times 1}$ and $\bm{f}_b^{v} \in \mathbb{R}^{H \times W \times 1}$ are the horizontal and vertical coordinate mapping matrix in $\bm{f}_b$, respectively, and $(u_{0},v_{0})$ is the pixel coordinate.

During training, the loss function is defined as the $L_1$ distance between the predicted warping flow $\bm{f}_b$ and its given ground truth $\bm{f}_{gt}$ as follows,
\begin{equation}
	\mathcal{L} = \left \| \bm{f}_{gt} - \bm{f}_b \right \|_1.
\end{equation}

\section{Dataset and Metrics}
To facilitate training and evaluation of the algorithms for unrestricted document image rectification, we first present the UDIR, 
a new dataset for unrestricted document image rectification.
Furthermore, we contribute the new quantitative metrics applicable for evaluating the rectification quality in terms of image similarity and distortion rectification.
In the following, we elaborate them separately.

\subsection{UDIR Dataset}\label{UDIR}
\textbf{Training Set.}
The training set is extended from the classic Doc3D dataset~\cite{9010747}.
It contains 100k distorted document images and their distortion-free ground truth.
In Doc3D~\cite{9010747}, each distorted image contains a complete document surrounded by a background region. 
To construct the training set for unrestricted document image rectification, we randomly crop such distorted document images to meet one of the following three conditions, including (a) with complete document boundaries, (b) with partial document boundaries, and (c) without any document boundaries, as illustrated in Fig.~\ref{fig:overview}.

\begin{figure}[t]
  \centering
  \includegraphics[width=1\linewidth]{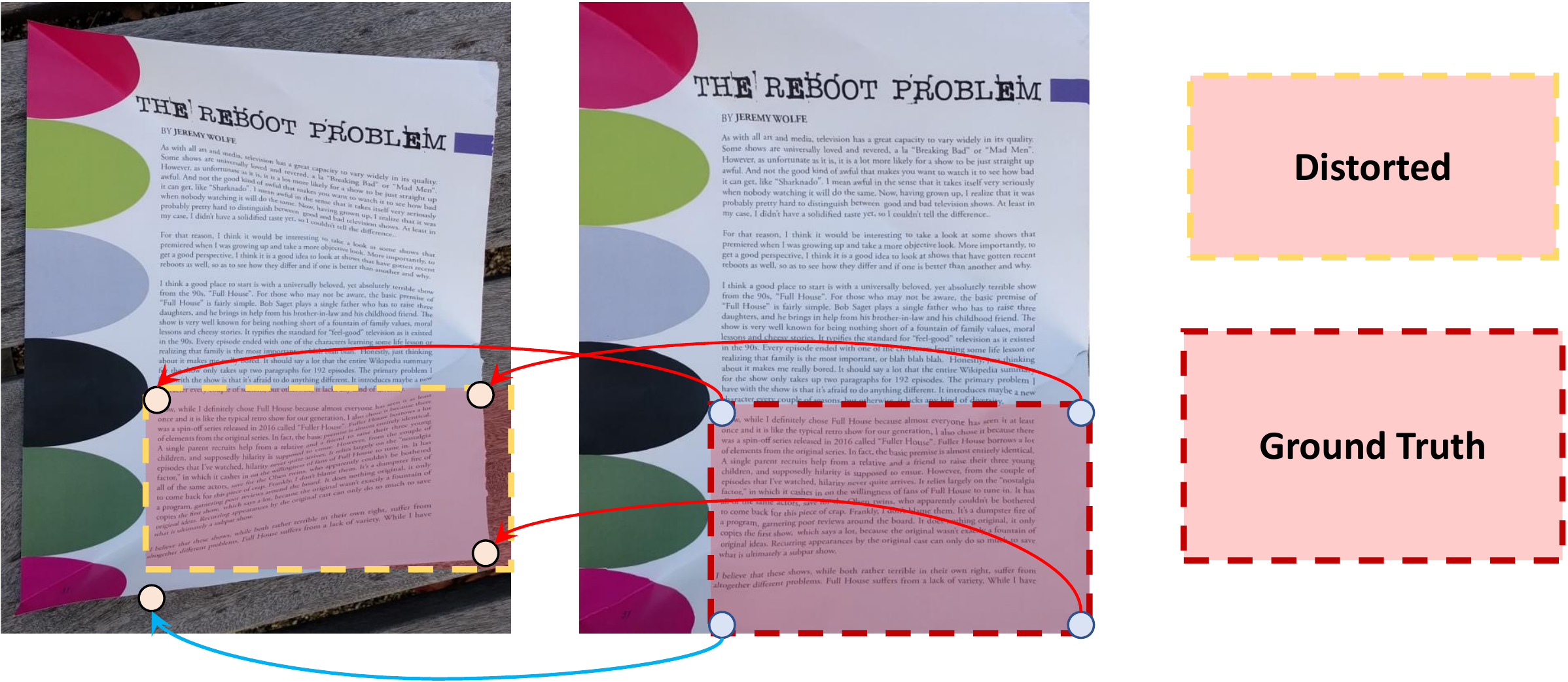}
  \caption{Illustration of the mapping relationship between the distorted image (with no boundaries) and the distortion-free one. The distortion-free image should cover the all content of the distorted one. Therefore, there are unexpected redundant regions that do not exist in the distorted one, highlighted with the blue arrow.
  The obtained warping flow is continuously distributed.}
  \label{fig:data_cons}
\end{figure}

\begin{figure*}[t]
  \centering
  \includegraphics[width=1\linewidth]{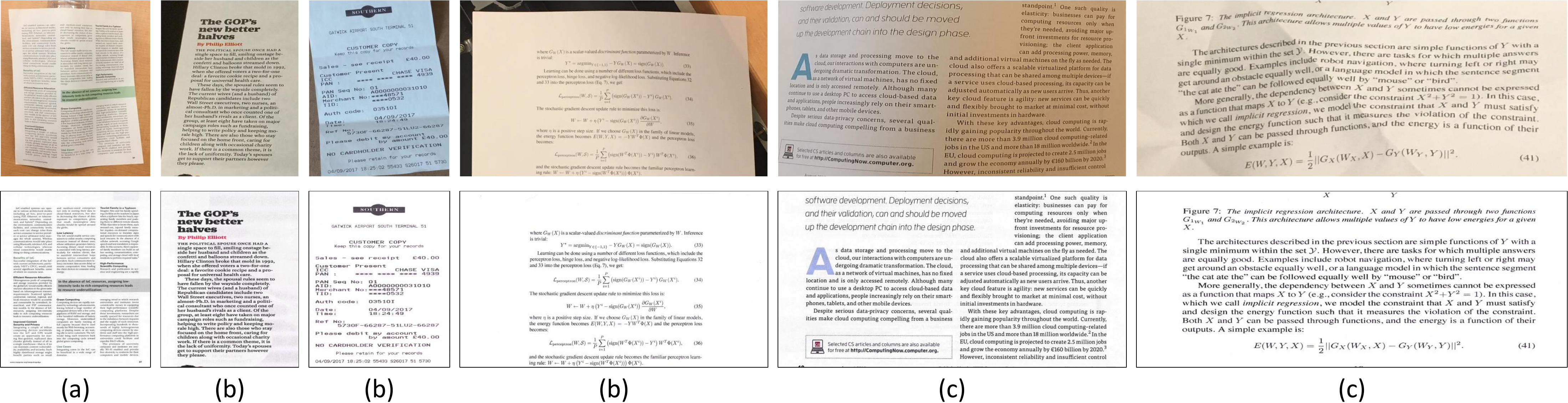}
  \caption{Sample images in the UDIR test set. The first and second rows show the distorted and ground truth images, respectively. (a), (b), and (c) represent the images with complete document boundaries, with partial document boundaries, and without any document boundaries, respectively.}
  \label{fig:test_set}
\end{figure*}

Meanwhile, to compose the backward warping flow as the ground truth for distortion rectification, we formulate the pixel-wise mapping relationship between each obtained distorted image and the corresponding distortion-free one,
as illustrated in Fig.~\ref{fig:data_cons}.
Note that the distortion-free image should cover the all content of the distorted one.
Therefore, there are unexpected redundant regions that do not exist in the distorted one (blue arrow in Fig.~\ref{fig:data_cons}).
The obtained flow is continuously distributed, but this does not affect its learning.
During the inference stage, the predicted flow pointing outside the input image directly fetches a fixed pixel value.
We set them as zero by default, which makes up the black regions in the rectified image (see the rectified images of DocTr++ in Fig.~\ref{fig:qua_eva_udoc}). 

\smallskip
\textbf{Test Set.}
Considering that there is no benchmark dataset for the evaluation of unrestricted document image rectification, we further construct a dataset that contains real-world camera-captured document images.
The dataset is extended from the classic DocUNet Benchmark dataset~\cite{8578592} that consists of 130 real document images and their ground truth.
Similarly, we manually crop such images to meet one of the following three conditions, including (a) with complete document boundaries, (b) with partial document boundaries, and (c) without any document boundaries.
The test set contains 195 images in total, and each of the three situations accounts for one-third.
Fig.~\ref{fig:test_set} presents some examples of our UDIR test set. We can see that the proposed test set covers the common distorted document images captured in daily life.

\subsection{Metrics}
In document image rectification, our input are genuine camera-captured distorted document images. Once rectified, these images still pertain to the category of natural scene images. On the contrary, our ground truth distortion-free document images are derived from the corresponding electronic PDF files and classified as computer-generated images~\cite{min2021screen}.
There are several existing works~\cite{zhai2020perceptual,min2020study} that have studied and summarized the quality assessment methods for these two types of images.
In contrast, in our evaluation, we aim to compute the discrepancies between these two entities.

In the existing literature on document image rectification~\cite{das2020intrinsic,9010747,das2021end,feng2021doctr,jiang2022revisiting,liu2020geometric,8578592,xie2021document,xie2020dewarping,xue2022fourier,feng2021docscanner,feng2022geometric,markovitz2020can,ma2022learning,zhang2022marior,li2019document}, the evaluations all calculate discrepancies across multiple dimensions between the rectified images and distortion-free ones. These dimensions include structural similarity of the images, pixel alignment, and edit distance of the recognized strings.
These distortion-free images are sourced from electronic PDF files, different from the rectified image obtained from the genuine camera-captured images. Consequently, this evaluation method might not be optimal, as the rectified images and the reference ones originate from different domains.
Numerous studies in the literature~\cite{min2017blind,min2018blind,min2017unified} have focused on blind quality assessment of natural scene images and screen content images.
Different from these images, 
document images exhibit distinct features, most notably the prevalence of numerous horizontal textlines.
Hence, future research could devise evaluation metrics based on the unique characteristics of document images, eliminating the need for reference images. 
For instance, one could detect textlines in the rectified images and quantify their curvature or deviation from an ideal horizontal orientation.

\begin{figure}[t]
  \centering
  \includegraphics[width=1\linewidth]{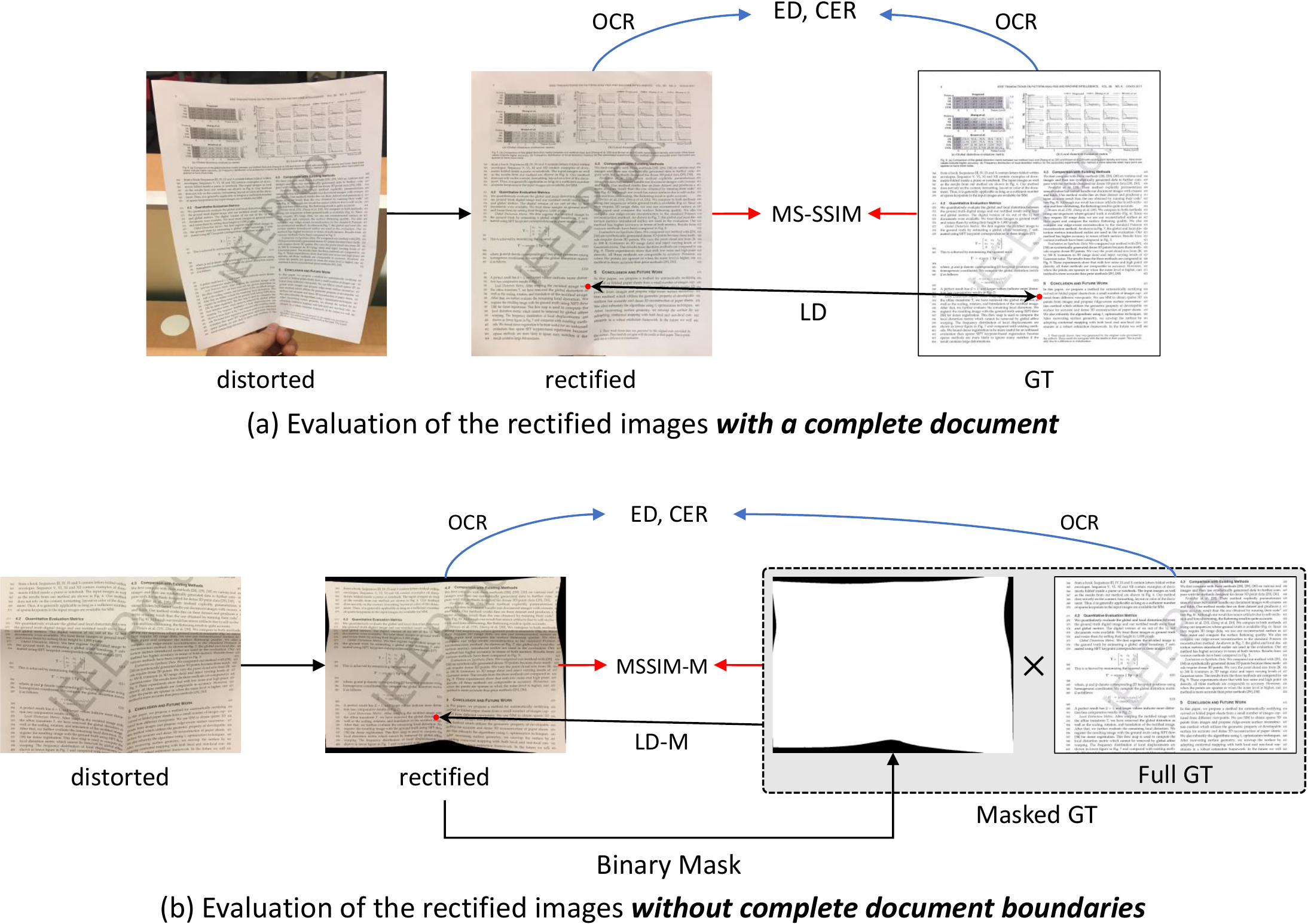}
  \caption{Illustration of the different ways for evaluating the rectified images (a) including a complete document and (b) without complete document boundaries. For the latter, to calculate the proposed MSSIM-M and LD-M, the original full GT image is first masked by the black regions in the rectified image. ``$\times$'' denotes pixel-wise multiplication.}
  \label{fig:metric}
\end{figure}

Fig.~\ref{fig:metric}(a) illustrates the evaluation way of typical metrics (\emph{i.e.}, MSSIM~\cite{1292216}, ED~\cite{levenshtein1966binary}, CER, and LD~\cite{Shaodi}) in the field of document image rectification, where the input distorted image should contain a complete document.
As shown in Fig.~\ref{fig:metric}(b), when the situation comes to the unrestricted document images, 
these metrics are no longer applicable, due to the unexpected black regions in rectified images.
Hence, we further propose metrics MSSIM-M and LD-M, described next.                                                                        

\smallskip
\textbf{MSSIM-M.} Structural SIMilarity (SSIM)~\cite{1284395} measures the per-patch similarity between two images. 
Since the perceivability of details rely on the sampling density of the image, Multi-scale Structural Similarity (MSSIM)~\cite{1292216} calculates the weighted summation of SSIM~\cite{1284395} across multiple scales. For document image rectification, all the rectified and ground truth flatbed-scanned images are first resized to a 598,400-pixel area, as recommended in DocUNet~\cite{8578592}.
Then, it is computed with a 5-level-pyramid,
following original MSSIM~\cite{1292216}. 
However, when the input distorted images do not contain complete documents, there will be some black regions in the rectified images, inconsistent with the ground truth one (see Fig.~\ref{fig:test_set}). To address the issue, we first mask the manually cropped ground truth image with these black regions and then calculate the MSSIM (see Fig.~\ref{fig:metric}(b)).
We term the obtained score as Multi-scale Structural SIMilarity Masked (MSSIM-M).

\smallskip
\textbf{LD-M.} Local distortion (LD)~\cite{Shaodi} first match the rectified image with the ground truth one using a dense SIFT-flow~\cite{5551153} $(\Delta \bm{x}, \Delta \bm{y})$, where $\Delta \bm{x}$ and $\Delta \bm{y}$ denote the horizontal and vertical displacement map of the matched pixels from the ground truth image to the rectified one, respectively.
LD is defined as the mean $L_2$ distance among all matched pixels, which measures the average local deformation of the rectified image.
Then, our Local Distortion Masked (LD-M) is calculated as the mean $L_2$ distance in the valid regions of the rectified image,
avoiding the invalid matches in the black regions.

\begin{figure*}[t]
  \centering
  \includegraphics[width=1\linewidth]{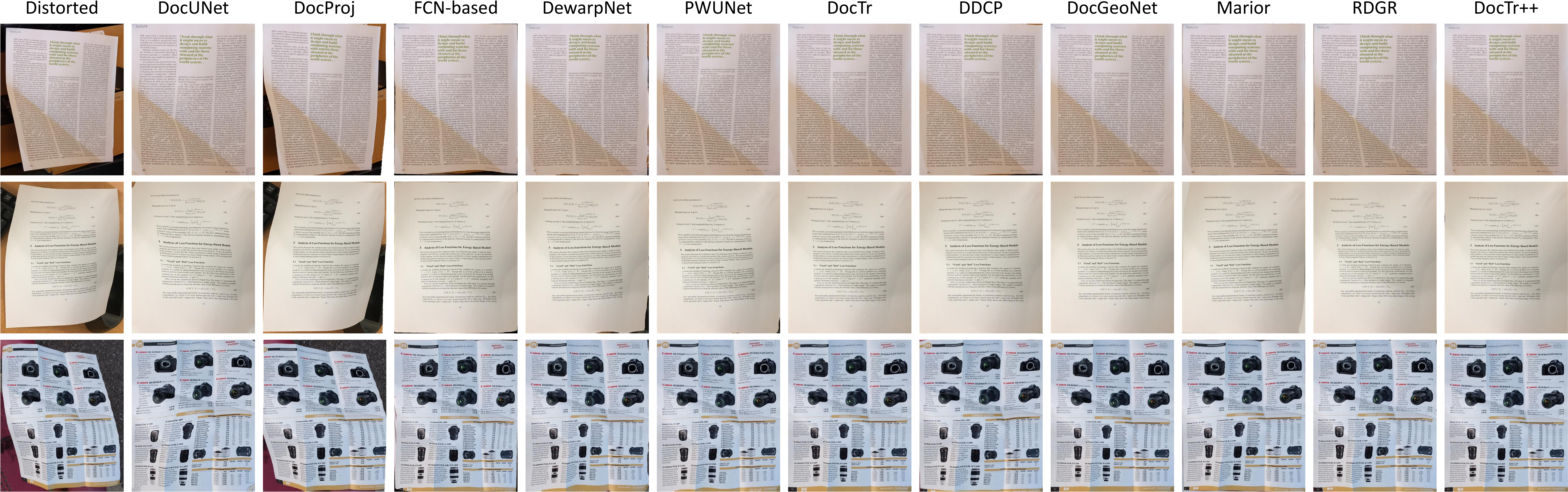}
  \caption{Qualitative comparisons with existing learning-based methods, including DocUNet~\cite{8578592}, DocProj~\cite{li2019document}, 
  FCN-based~\cite{xie2020dewarping}, DewarpNet~\cite{9010747}, PWUNet~\cite{das2021end}, DocTr~\cite{feng2021doctr}, DDCP~\cite{xie2021document}, DocGeoNet~\cite{feng2022geometric}, Marior~\cite{zhang2022marior}, RDGR~\cite{jiang2022revisiting}, and our DocTr++, from left to right. Zoom in for a better view.}
  \label{fig:qua_eva1}
\end{figure*}

\smallskip
\textbf{ED and CER.} Edit Distance (ED)~\cite{levenshtein1966binary} measures how dissimilar two strings are to one another, which is defined as the minimum number of operations required to transform one string into the reference one. It can be efficiently computed based on the dynamic programming algorithm. Specifically, the involved operations include deletions $(d)$, insertions $(i)$, and substitutions $(s)$. Then, Character Error Rate (CER) can be computed as follows,
\begin{equation}\label{equ:cer}
CER=(d+i+s)/{N_c} ,
\end{equation}
where $N_c$ is the character number of the reference string. CER depicts the percentage of characters in the reference text incorrectly recognized in the rectified image. The lower the CER value (with 0 being a perfect score), the better the performance of the rectification method.
As shown in Fig.~\ref{fig:metric}, since $N_c$ is a constant for the full GT distortion-free image, 
here we do not mask it with the rectified black regions,
which does not affect the evaluation of OCR performance.

\setlength{\tabcolsep}{1.6mm}
\begin{table}[t]
\centering
	\caption{Quantitative comparisons with existing methods on the DocUNet Benchmark dataset~\cite{8578592}. ``$\uparrow$'' indicates the higher the better and ``$\downarrow$'' means the opposite. The red and blue highlight the highest and second-highest performance, respectively.}
\begin{tabular}{l|c|cccc} 
   \toprule
   \textbf{Methods} & \textbf{Venue} &\textbf{MSSIM} $\uparrow$ &\textbf{LD} $\downarrow$  &\textbf{ED} $\downarrow$ &\textbf{CER} $\downarrow$   \\ 
   \midrule
    Distorted & / & 0.25 & 20.51 & 1552.22 & 0.5089 \\
            \midrule
    DocUNet~\cite{8578592} & \emph{CVPR'18} & 0.41 & 14.19 & 1259.83 & 0.3966 \\
    DocProj~\cite{li2019document} & \emph{TOG'19} & 0.29 & 18.01  & 1165.93 & 0.3818 \\

    FCN-based~\cite{xie2020dewarping} & \emph{DAS'20}  & 0.45 & 7.84 & 1031.40 & 0.3156 \\ 

    DewarpNet~\cite{9010747} & \emph{ICCV'19}  & 0.47 & 8.39  & 525.45 & 0.2102 \\

    DocTr~\cite{feng2021doctr} & \emph{MM'21} & \textcolor{red}{\textbf{0.51}} & 7.76  & 464.83 & 0.1746 \\
    
    PWUNet~\cite{das2021end} & \emph{ICCV'21} & 0.49 & 8.64  & 743.32 & 0.2623 \\    
    
    DDCP~\cite{xie2021document}  & \emph{ICDAR'21} & 0.47 & 8.99 & 745.35 & 0.2626 \\

    DocGeoNet~\cite{feng2022geometric} & \emph{ECCV'22} & \textcolor{blue}{\textbf{0.50}} & 7.71 & \textcolor{blue}{\textbf{379.00}} & \textcolor{red}{\textbf{0.1509}} \\ 

    Marior~\cite{zhang2022marior} & \emph{MM'22} & 0.48 & \textcolor{red}{\textbf{7.44}} & 593.80 & 0.2136 \\

    RDGR~\cite{jiang2022revisiting} & \emph{CVPR'22} & \textcolor{blue}{\textbf{0.50}} & 8.51 & \textcolor{red}{\textbf{420.25}} & \textcolor{blue}{\textbf{0.1559}}  \\
        \midrule
        DocTr++ & / & \textcolor{red}{\textbf{0.51}}  & \textcolor{blue}{\textbf{7.52}} & 447.47 & 0.1695  \\
   \bottomrule
\end{tabular}
\label{tab:t1}
\end{table} 

\section{Experiments}
 
\subsection{Implementation Detail}
\textbf{Inference Way.}
Following DocTr~\cite{feng2021doctr}, during training, the image size $(H, W)$ is set as $(288, 288)$.
For evaluation, the real-world distorted document images of arbitrary resolutions are first resized to $(288, 288)$. The obtained images are then fed into the network that produces a warping flow of shape $(288, 288)$. Finally, we resize the flow map to the original shape of the input distorted image and unwarp it with the obtained flow map. In this way, we produce the output high-resolution rectified document image with legible text.

\smallskip
\textbf{Training.}
We use the AdamW optimizer~\cite{loshchilov2017decoupled} and OneCycle learning rate schedule~\cite{smith2019super} with a maximum learning rate of $10^{-4}$.
The epochs for the warm-up phase account for $10\%$.
The network is trained for 65 epochs with a batch size of 12.
We employ 4 NVIDIA GeForce RTX 2080Ti GPUs for training.
Besides, to generalize better to real data with complex illumination conditions, we add a jitter in the HSV color space to magnify illumination and document color variations. 

\begin{figure*}[t]
  \centering
  \includegraphics[width=1\linewidth]{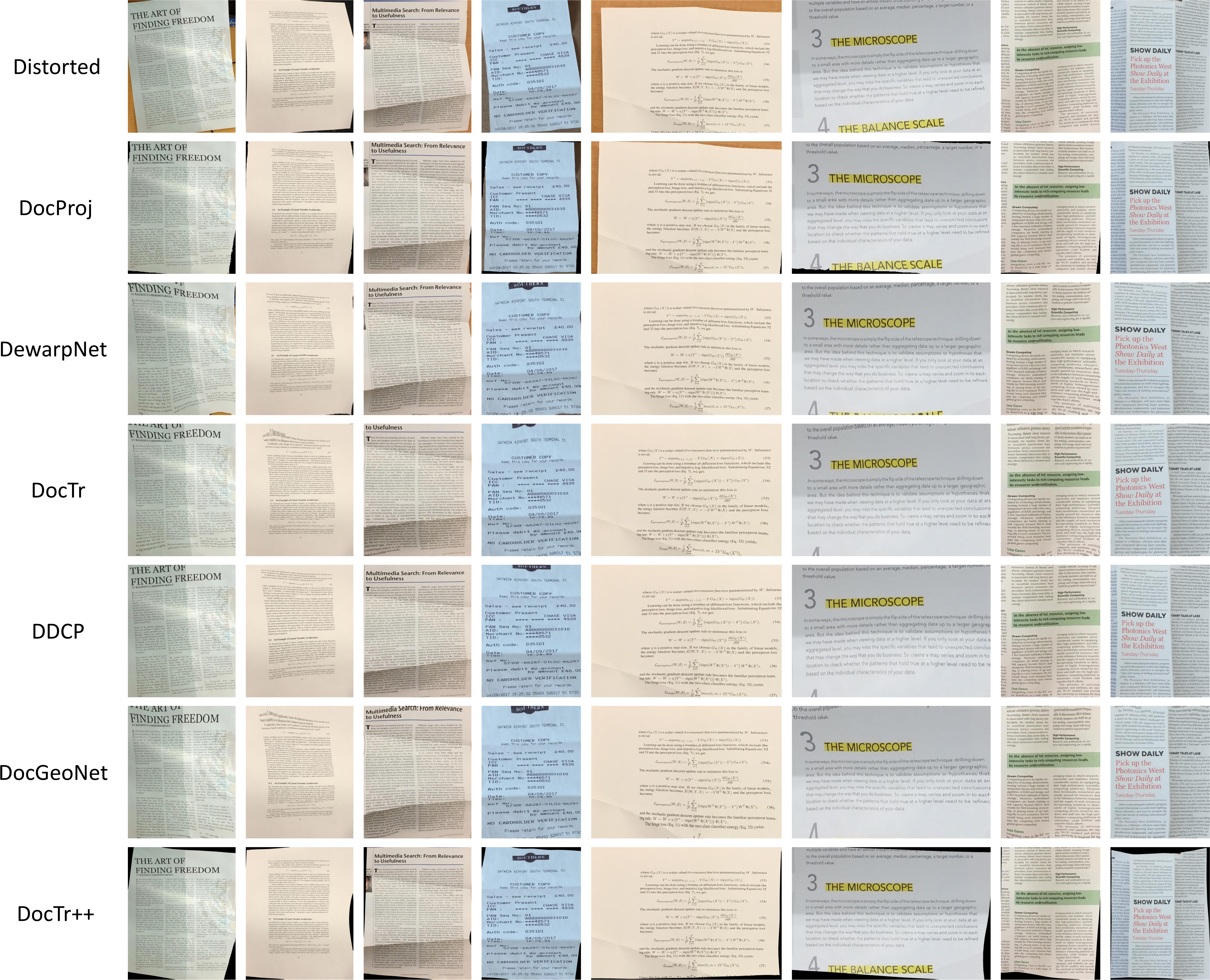}
  \caption{Qualitative results of DocTr++ and other methods on the UDIR test set. The first row shows the input distorted document images, and the other rows are rectified results of the corresponding methods. The results of our DocTr++ present less distortion than the other methods.}
  \label{fig:qua_eva_udoc}
\end{figure*}

\setlength{\tabcolsep}{1.15mm}
\begin{table}[t]
\centering
	\caption{Quantitative comparisons with existing methods on the proposed UDIR test set. ``$\uparrow$'' indicates the higher the better and ``$\downarrow$'' means the opposite. The red and blue highlight the highest and second-highest performance, respectively.}
\begin{tabular}{l|c|cccc} 
   \toprule
   \textbf{Methods} & \textbf{Venue} &\textbf{MSSIM-M} $\uparrow$ &\textbf{LD-M} $\downarrow$  &\textbf{ED} $\downarrow$ &\textbf{CER} $\downarrow$   \\ 
   \midrule
    Distorted & / & 0.31 & 18.87 & 1181.47 & 0.3986 \\
    \midrule
    DocProj~\cite{li2019document} & \emph{TOG'19} &   0.31 & 19.11 & 932.11  &   0.2995 \\

    DewarpNet~\cite{9010747} & \emph{ICCV'19}  &   0.36 & \textcolor{blue}{\textbf{17.91}}  & 1037.04  &  0.3454   \\

    DocTr~\cite{feng2021doctr} & \emph{MM'21} &  \textcolor{blue}{\textbf{0.38}}  & 18.80 & \textcolor{blue}{\textbf{840.67}}    &   \textcolor{blue}{\textbf{0.2988}}  \\
    
    DDCP~\cite{xie2021document}  & \emph{ICDAR'21} &  0.36 & 19.26 & 984.63  &  0.3324  \\  

    DocGeoNet~\cite{feng2022geometric} & \emph{ECCV'22} &  \textcolor{blue}{\textbf{0.38}} & 18.60 & 872.69  &  0.3024     \\ 

    \midrule
    DocTr++ & / &   \textcolor{red}{\textbf{0.45}}   &   \textcolor{red}{\textbf{12.47}}  
    &  \textcolor{red}{\textbf{666.49}}    &  \textcolor{red}{\textbf{0.2288}}      \\
   \bottomrule
\end{tabular}
\label{tab:t2}
\end{table}

\setlength{\tabcolsep}{2.7mm}
\begin{table}[h]
	\caption{Quantitative comparisons of the existing learning-based methods in terms of running efficiency on the DocUNet Benchmark dataset~\cite{8578592}.
    ``$\uparrow$'' indicates the higher the better.}
	\centering
	\begin{tabular}{c|c|c|c}  
		\toprule
		\textbf{Methods} & \textbf{Venue} & \textbf{FPS} $\uparrow$ & \textbf{Parameters (M)} \\  
		\midrule
		DocUNet~\cite{8578592} & \emph{CVPR'18} & 0.21 & 58.6 \\
		DocProj~\cite{li2019document} & \emph{TOG'19} & 0.11 & 47.8 \\ 
		DewarpNet~\cite{9010747} & \emph{ICCV'19} & 7.14 & 86.9 \\  
 		FCN-based~\cite{xie2020dewarping} & \emph{DAS'20} & 1.49 & \textbf{23.6} \\
		PWUNet~\cite{das2021end} & \emph{ICCV'21} & - & - \\  
		DocTr~\cite{feng2021doctr} & \emph{ACM MM'21} & 7.40 & 26.9 \\ 
		DDCP~\cite{xie2021document} & \emph{ICDAR'22} & - & - \\ 
		DocGeoNet~\cite{feng2022geometric} &  \emph{ECCV'22} & 7.82 & 24.8 \\ 
		\midrule
		DocTr++ & / & \textbf{11.29} & 26.6 \\ 
		\bottomrule
	\end{tabular}
	\label{com:eff}
\end{table}
\smallskip
\textbf{Evaluation.}
Firstly, to calculate MSSIM and MSSIM-M, the weight for each level of the image pyramid is set as 0.0448, 0.2856, 0.3001, 0.2363, and 0.1333, following methods~\cite{feng2021doctr,feng2021docscanner,feng2022geometric}.
The Matlab version is 2019a.
Secondly,
we use Tesseract (v5.0.1)~\cite{4376991} as the OCR engine to recognize the text string of the rectified image and the ground truth one, as recommended in previous works~\cite{feng2021doctr,feng2021docscanner,feng2022geometric}.
For the DocUNet Benchmark dataset~\cite{8578592}, we select 60 images for OCR evaluation, following DocTr~\cite{feng2021doctr}.
For our UDIR test set, we select 70 images.
Note that within these two sub-datasets, the text present in all document images is exclusively in English.
Besides, in such selected images, the text should make up the majority of the content. This is because if the text is rare in an image, the character number ${N_c}$ in Equation~\eqref{equ:cer} is a small number, leading to a significant variance for CER.

Notably, we observe that the $127^{th}$ and $128^{th}$ distorted document images in the DocUNet Benchmark dataset~\cite{8578592} are rotated by 180 degrees,
which do not match the ground truth documents.
This inconsistency is ignored by our DocTr~\cite{feng2021doctr} and many previously published methods~\cite{9010747,das2021end,feng2021doctr,jiang2022revisiting,liu2020geometric,8578592,xie2021document,xie2020dewarping,xue2022fourier,markovitz2020can,ma2022learning,zhang2022marior,li2019document}.
In this paper,
the reported performances of our DocTr++ and other methods are based on the corrected dataset.

\subsection{Experimental Results}
We quantitatively and qualitatively evaluate the proposed DocTr++ with existing learning-based methods.
The evaluation is conducted on the widely-used DocUNet Benchmark dataset~\cite{8578592} and the proposed UDIR test set.
Note that different from the former dataset, the latter one contains various images with complete or incomplete document boundaries.

\begin{figure*}[t]
  \centering
  \includegraphics[width=1\linewidth]{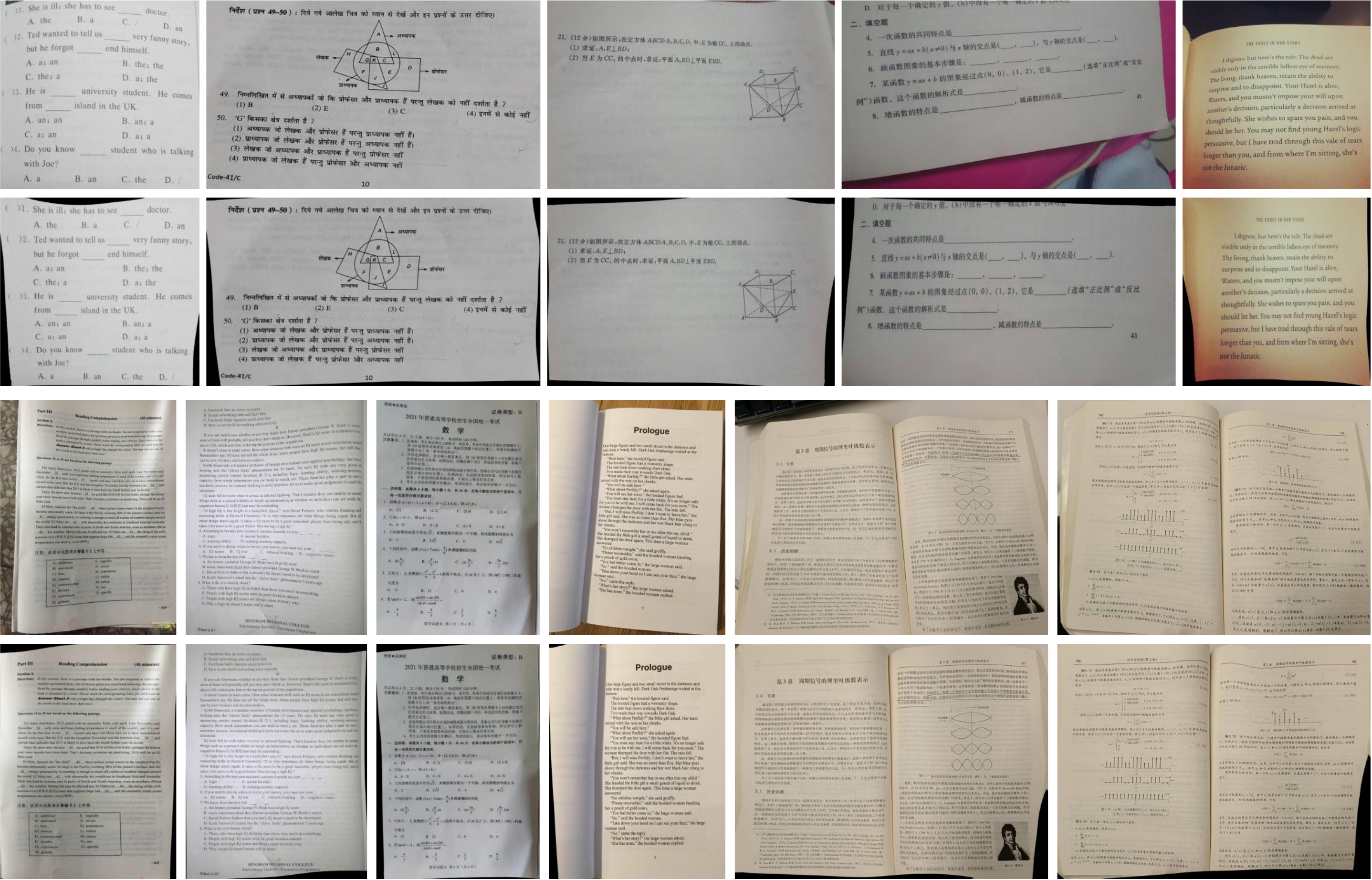}
  \caption{Qualitative results of DocTr++ on some common real-world distorted document images, including test papers, book pages, and text paragraphs. The first and third rows display the input distorted images, while the second and fourth rows present their corresponding rectified results.}
  \label{fig:qua_eva2}
\end{figure*}

\begin{figure*}[t]
  \centering
  \includegraphics[width=1\linewidth]{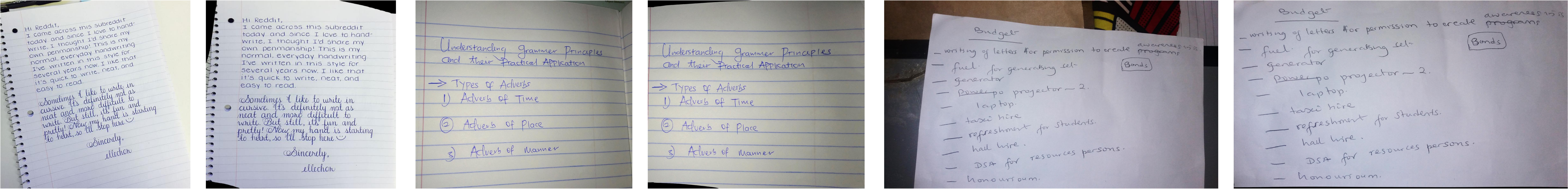}
  \caption{Qualitative results of DocTr++ on three real-world distorted handwriting images. For each sample, we show the distorted and rectified images, from left to right. Such distorted document images do not contain complete document boundaries. They are effectively restored and the rectified textlines are straight along the horizontal direction.
  }
  \label{fig:hand}
\end{figure*}

\smallskip
\textbf{Quantitative Comparison.}
We first compare DocTr++ with the existing learning-based methods on the DocUNet Benchmark dataset~\cite{8578592}. As mentioned above, all the distorted images in this dataset incorporate complete documents. Following prior methods~\cite{feng2021doctr,feng2022geometric,feng2021docscanner,xie2020dewarping,zhang2022marior}, we add a document localization module to remove the noisy backgrounds around the document before the distortion rectification.
Here we adopt the same pre-possessing module~\cite{Qin_2020} used in DocTr~\cite{feng2021doctr}.
The comparison results are shown in Table~\ref{tab:t1}.
Compared with DocTr~\cite{feng2021doctr},
DocTr++ improves LD and CER from 7.76 to 7.52 and $17.46\%$ to $16.95\%$, respectively.
Moreover, our DocTr++ also shows comparable performance with the state-of-the-art RDGR~\cite{jiang2022revisiting} and outperforms the recent Marior~\cite{zhang2022marior} on metric CER by $4.41\%$.

\begin{figure}[t]
  \centering
  \includegraphics[width=0.93\linewidth]{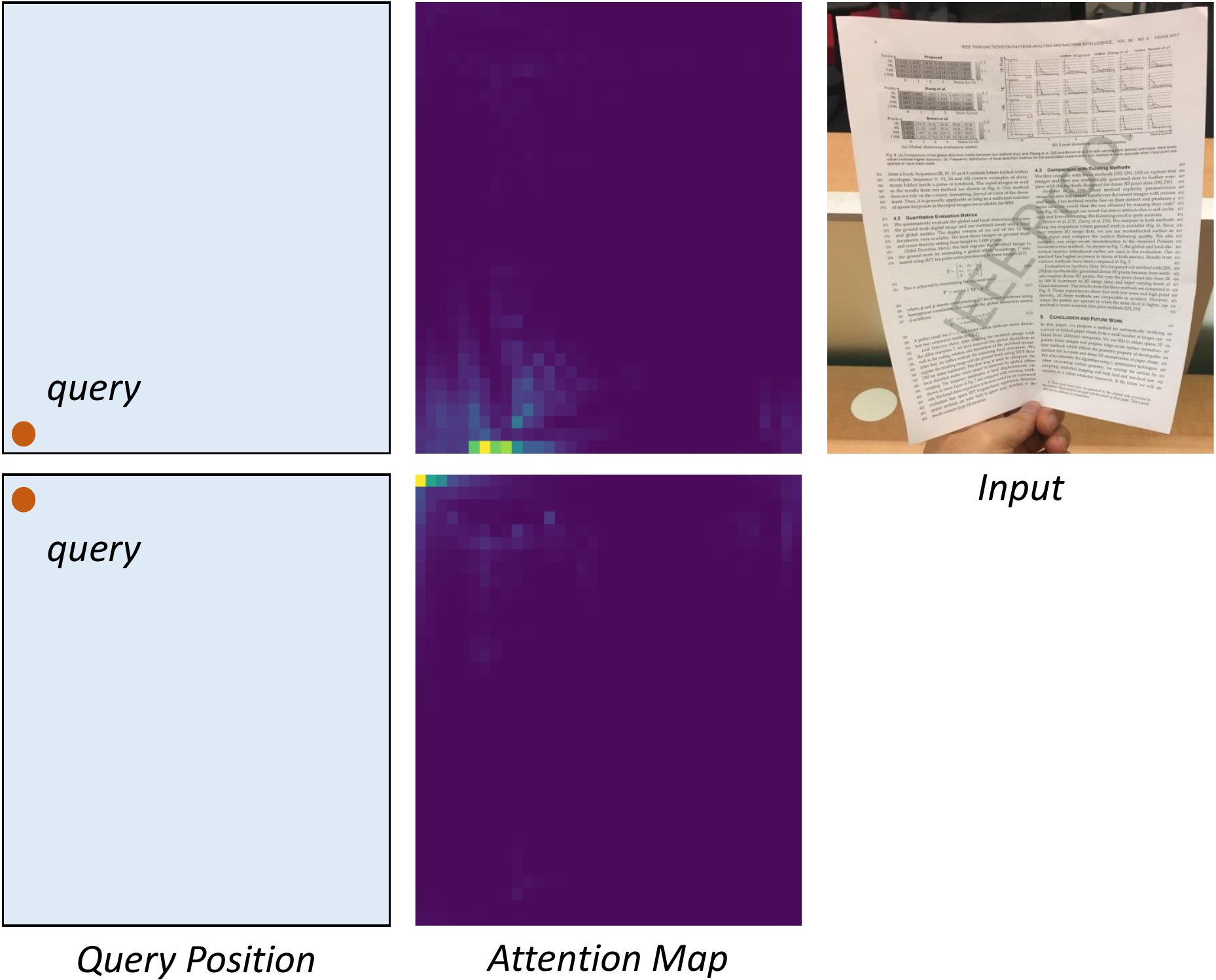}
  \caption{Attention map visualization of the learnable query in the rectification decoder. We take the attention maps from its last block. As we can see, each query is responsible for the rectification of a specific region in the input distorted image that is accurately attended. 
  The visualization effectively illustrates the role of the learnable queries in our method.}
  \label{fig:attention}
\end{figure}

Table~\ref{tab:t2} summarizes the results of our DocTr++ and the previously published methods on the proposed UDIR test set.
Note that here we only compare the methods with released codes.
As we can see,
our DocTr++ performs better than these methods on all metrics.
Notably, we outperform DocTr on MSSIM-M by $7\%$ and yield a relative improvement on LD-M by $33.67\%$.
Such results indicate the restoration ability of DocTr++ on processing unrestricted document images.

\smallskip
\textbf{Qualitative Comparison.}
To clearly show the effectiveness of our DocTr++, we carry out qualitative comparisons with other existing methods to provide visual evidence.

Firstly, in Fig.~\ref{fig:qua_eva1}, we qualitatively compare the rectified results of the existing learning-based methods and our DocTr++ on the DocUNet Benchmark dataset~\cite{8578592}.
As we can see, our DocTr++ reveals superior rectification performance over other methods.
Secondly, as shown in Fig.~\ref{fig:qua_eva_udoc}, we present the rectified results of other learning-based methods and our DocTr++ on the proposed UDIR test set. As can be seen, our method can effectively rectify the unrestricted distorted document images including incomplete documents or local text paragraphs.
Besides, compared with these previous methods, our rectified images reveal less distortion.
These results demonstrate the strong generation ability of our proposed method in restoring various real-world distorted document images.

\begin{figure}[t]
  \centering
  \includegraphics[width=0.98\linewidth]{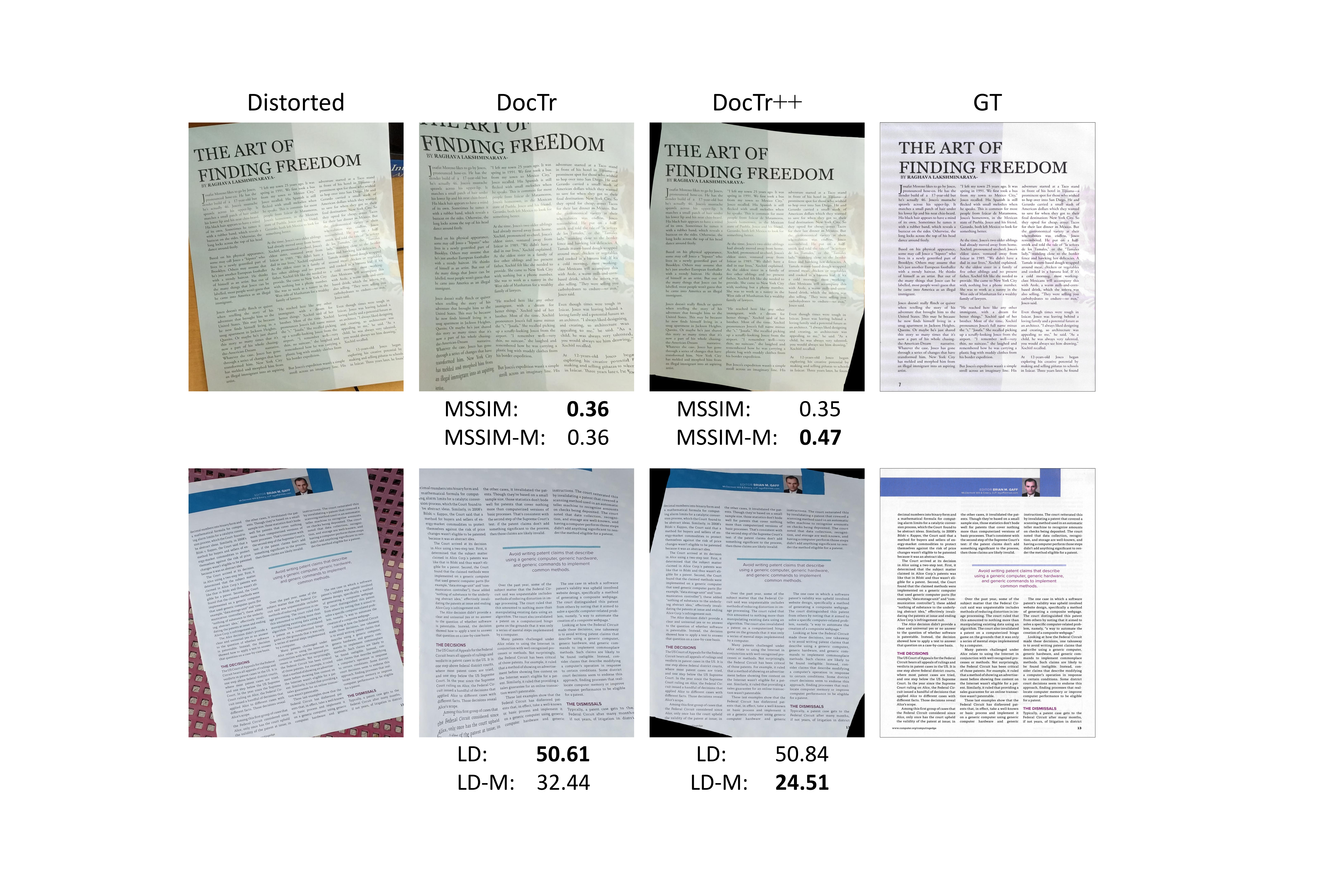}
  \caption{Failure cases of metric MSSIM and LD. Compared with DocTr~\cite{feng2021doctr}, the rectified images of DocTr++ show less distortion. However, the metric MSSIM and LD can not well perceive their difference, while the proposed MSSIM-M and LD-M are more robust and reliable.}
  \label{fig:metric_aba}
\end{figure}

\begin{figure*}[t]
  \centering
  \includegraphics[width=0.8\linewidth]{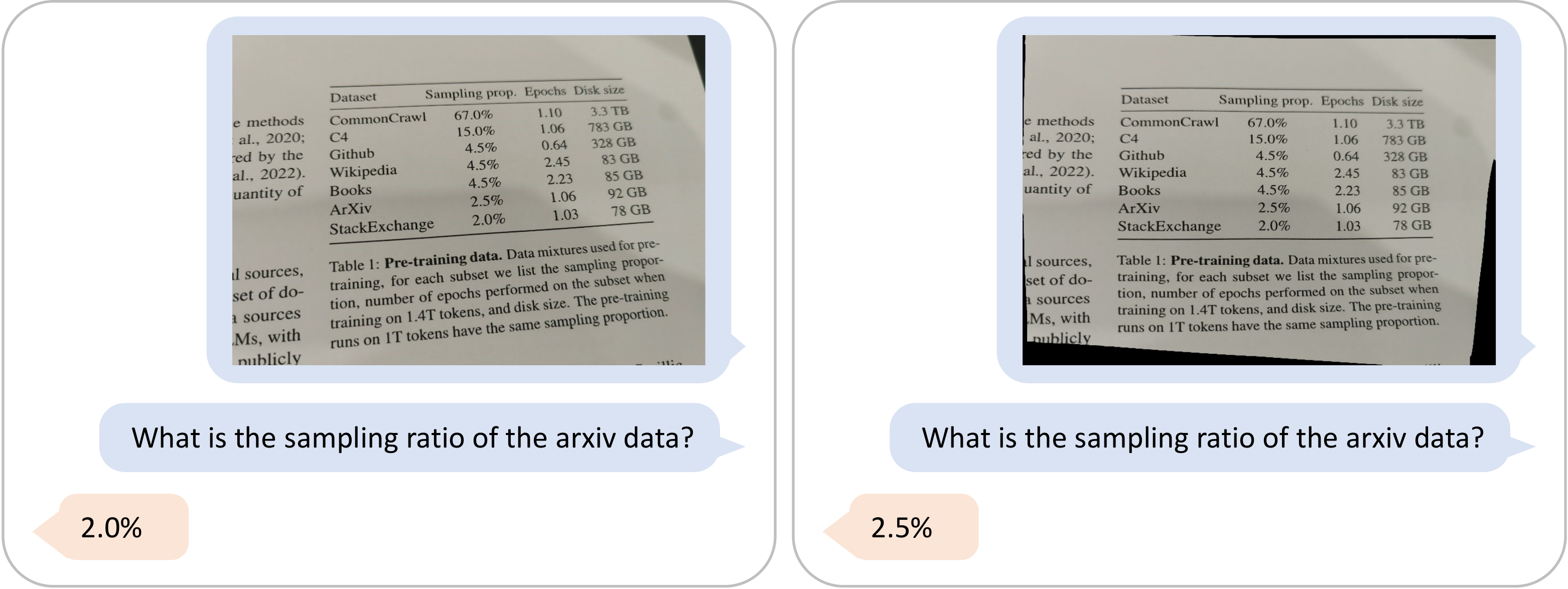}
  \caption{Example to demonstrate the impact of geometric correction on document understanding based on large multimodal models(LMMs). The left and right respectively show the answers to the same instruction before and after correction. The employed LMM is Qwen-VL~\cite{Qwen-VL}.}
  \label{fig:under_case}
\end{figure*}

\setlength{\tabcolsep}{1.6mm}
\begin{table}[t]
\centering
	\caption{Ablations about the network architecture of DocTr++ on the proposed UDIR test set. ``$\uparrow$'' indicates the higher the better and ``$\downarrow$'' means the opposite. Settings used in our final model are underlined.}
\begin{tabular}{cccccc} 
   \toprule
   \textbf{Methods} & \textbf{Setting} &\textbf{MSSIM-M} $\uparrow$ &\textbf{LD-M} $\downarrow$  &\textbf{ED} $\downarrow$ &\textbf{CER} $\downarrow$   \\

   \midrule 
   \multirow{2}*{Encoder} & \underline{w/} &  \textbf{0.45}    &    \textbf{12.47}   &   \textbf{666.49}     &   \textbf{0.2288}   \\
   & w/o  &  0.44  & \textbf{12.47} & 680.44  & 0.2353  \\
   
   \midrule 
   \multirow{2}*{Decoder} & \underline{w/} &  \textbf{0.45}    &    \textbf{12.47}   &   \textbf{666.49}    &   \textbf{0.2288}   \\
   & w/o  &  \textbf{0.45}   &  12.56  &  684.67 &  0.2293     \\

   \midrule 
   \multirow{2}*{Query} & \underline{learned}  &  \textbf{0.45}    &    \textbf{12.47}   &   \textbf{666.49}    &   \textbf{0.2288}   \\
   & fixed &   \textbf{0.45}    &   12.52 &  717.81 &   0.2408   \\
   
   \midrule 
   \multirow{2}*{Upsample} & \underline{learned} &  \textbf{0.45}    &    \textbf{12.47}   &   \textbf{666.49}    &   \textbf{0.2288}   \\
   & bilinear &   \textbf{0.45}    &   12.77 &  689.97 &   0.2375    \\

   \midrule 
   \multirow{2}*{Flow} & \underline{continuous}  &  \textbf{0.45}    &    \textbf{12.47}   &   \textbf{666.49}     &   \textbf{0.2288}   \\
   & discontinuous & 0.44 & 12.96 & 715.43 & 0.2421 \\

   \bottomrule
\end{tabular}
\label{tab:t3}
\end{table}

\smallskip
\textbf{Efficiency Comparison.}
Furthermore, we compare the running efficiency and the parameter numbers of the state-of-the-art solutions and our DocTr++.
The results are summarized in Table~\ref{com:eff}.
As we can see,
compared with the previous DocTr~\cite{feng2021doctr}, our enhanced DocTr++ has slightly more parameters, but a much higher running efficiency,
This is attributed to the hierarchical architecture of DocTr++ that effectively reduces the $O(N^2)$ complexity of the self-attention process~\cite{Vaswani2017AttentionIA}.
Additionally,
in terms of the FPS (frames per second) and the model parameter size, our DocTr++ is also competitive compared with the state-of-the-art methods.

\subsection{Ablation Study}
We further conduct extensive quantitative and qualitative experiments to validate each component of our DocTr++.

\smallskip
\textbf{Architecture Setting.}
Table~\ref{tab:t3} studies the architecture setting of DocTr++.
Without either the distortion encoder or the rectification decoder,
the performances become worse.
It should be noted that in both experiments, the number of the transformer encoder or decoder layer is set as 12, equal to our DocTr++.
The results demonstrate the significance of a strong encoder and decoder for distortion representation learning and distortion correction prediction, respectively.
Then,
we study the learnable query with a fixed one.
As we can see,
performances are better with the learnable query.
Next,
we validate the upsample module~\cite{feng2021doctr} for warping flow prediction.
We can see that, compared with the bilinear upsample module, the learnable one produces better performance.
The likely reason is that the learnable way is capable of rectifying finer deformations.
As mentioned in Sec.~\ref{UDIR}, our ground truth warping flow is continuously distributed.
To ablate it, we set the flow that points to the outside of the input distorted image as value -1, following DocUNet~\cite{8578592}.
The results show that learning the continuously distributed flow map is more effective and yields better performance.

\begin{figure}[t]
  \centering
  \includegraphics[width=1\linewidth]{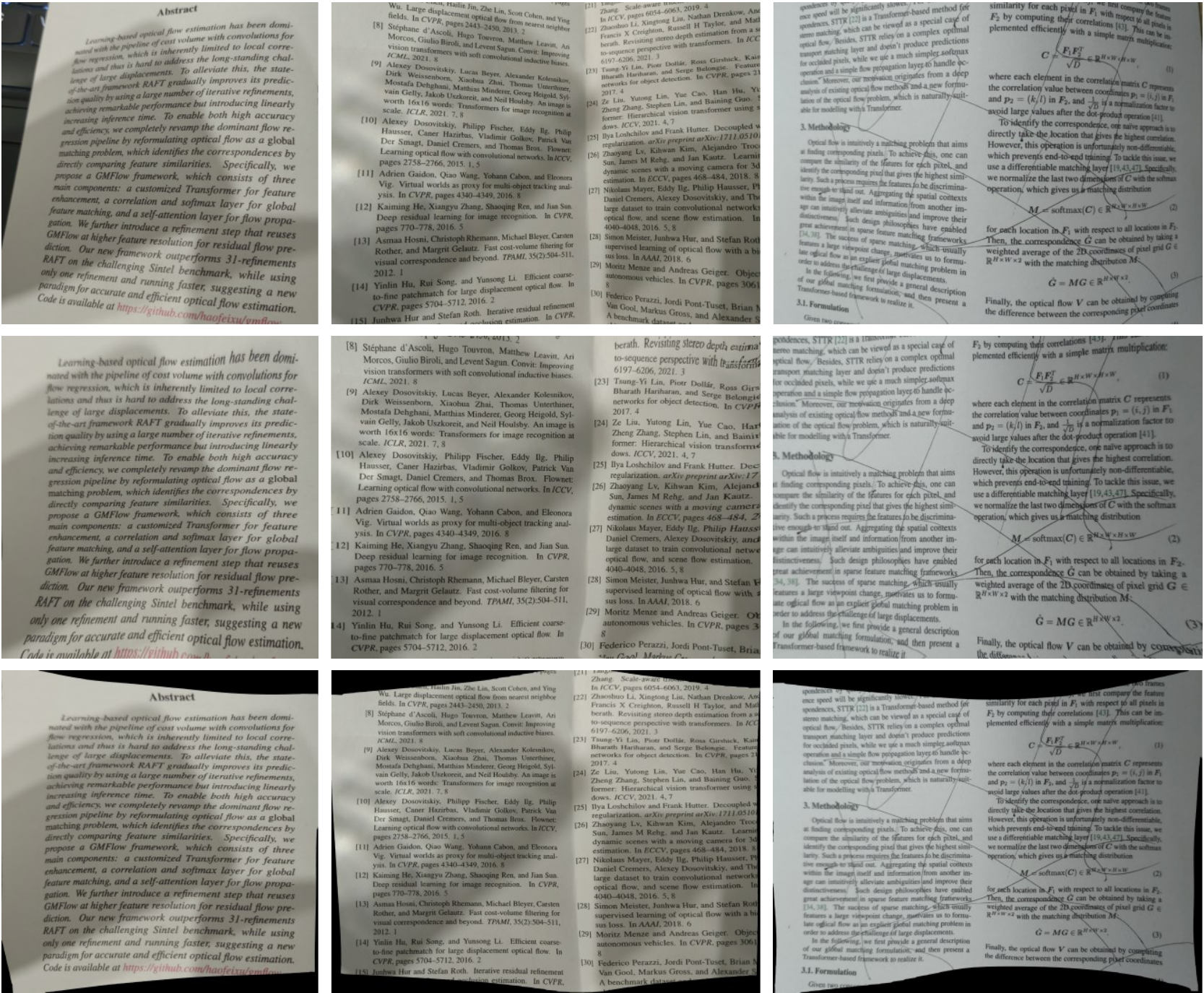}
  \caption{Rectification results of DocTr~\cite{feng2021doctr} and DocTr++ on distorted document images with shadows or noise. The first row presents the input distorted document images, the second row displays the corrected images by DocTr~\cite{feng2021doctr}, and the bottom row showcases the rectified images by our DocTr++. Notably, samples in the first and second columns contain shadows, while the sample in the third column exhibits noise. Due to the inability of other methods to handle images with absent or incomplete document boundaries, we present only the rectification results of DocTr~\cite{feng2021doctr}.}
  \label{fig:noise}
\end{figure}

To better illustrate the role of the query that is the input to the rectification decoder, we visualize the attention map of some typical queries.
As shown in Fig.~\ref{fig:attention},
each query is responsible for rectifying a specific region in the input distorted image that is accurately attended.
Such explainable results explain how our decoder works.
This mechanism is similar to the classic detector DETR~\cite{carion2020end},
where the object queries are used to decode the boxes with specific patterns. 

\smallskip
\textbf{Robustness of Proposed Metrics.}
In Fig.~\ref{fig:metric_aba},
we show two failure cases of the metric MSSIM and LD in evaluating the rectified unrestricted document images.
As we can see, compared with DocTr~\cite{feng2021doctr}, 
the rectified images of DocTr++ are less distorted. 
Nevertheless,
MSSIM and LD cannot effectively reveal their difference,
due to the black area existing in the rectified image.
In contrast, 
the proposed MSSIM-M and LD-M are more robust and reliable.

\smallskip
\textbf{Other Results.}
To better evaluate our method, we collect some other real-world distorted document images, including test papers, book pages, and text paragraphs. The rectified results are displayed in Fig.~\ref{fig:qua_eva2}. As we can see, such distorted images common in daily life are well rectified by our method.

Besides, as shown in Fig.~\ref{fig:hand}, we also present some handwriting paper cases. 
It is evident that our method produces neat rectified images with straight textlines, demonstrating its strong generalization ability and robustness.

As present in Fig.~\ref{fig:noise}, we further present the correction results of DocTr~\cite{feng2021doctr} and DocTr++ for processing distorted document images with shadows or dirt/noise.
Owing to the limitations of the existing methods in handling distorted images without document boundaries or with incomplete document boundaries, here we only present the correction results of DocTr~\cite{feng2021doctr}.
The correction results of other existing methods are akin to those of DocTr~\cite{feng2021doctr}.
It can be observed that our DocTr++ is robust against these complex shadows and messy noise strokes, consistently delivering stable correction results.
These results demonstrate the robustness and strong generalizability of our DocTr++.

\smallskip
\textbf{Impact on Downstream Tasks.}
Our DocTr++ can serve as a preprocessing step for various visual or multimodal understanding tasks, enhancing the performance of these tasks.
Firstly, as evident in Fig.~\ref{fig:doc_class} (bottom row),
the detection rate of text significantly improves after applying our model for distortion rectification.
Secondly, as shown in Table~\ref{tab:t1} and Table~\ref{tab:t2}, the character error rates (CER) on the DocUNet~\cite{8578592} and our UDIR benchmark datasets were 50.89$\%$ and 39.86$\%$, respectively. 
After employing our model for distortion correction, these character error rates were reduced to 16.95$\%$ and 22.88$\%$. 
These qualitative and quantitative results underscore the efficacy of geometric correction in improving the performance of OCR tasks.
Thirdly, we explored the influence of DocTr++ on the multimodal understanding abilities of OCR-free large multimodal models (LMMs)~\cite{liu2023llava}.
The employed LMM is Qwen-VL~\cite{Qwen-VL}.
As shown in Fig.~\ref{fig:under_case}, after undergoing geometric distortion correction with our DocTr++, Qwen-VL~\cite{Qwen-VL} is able to accurately give the response.

\section{Limitation Discussion}
For document image rectification,
it is advantageous to extract the cues from the 
distorted document boundaries and curved textlines that contain the structural information about the page deformation. 
In particular, for the images without complete document boundaries,
it is helpful to exploit the curved textlines in the distorted image that are straight in the rectified one.
In this work, our method implicitly encodes such representations, without explicitly considering the geometrical constraints that bridge the distorted and the rectified document images.
In future work, we will further explore the potential of these attributes for improving the quality and accuracy of rectified document images.

\section{Conclusion}
In this work, we introduce DocTr++, a novel deep network for document image rectification.
Our method breaks through the scenario limitations of existing rectification methods and is capable of restoring various distorted document images commonly encountered in everyday life.
To achieve superior rectification results, we enhance the original DocTr with a hierarchical structure consisting of a distortion encoder and decoder. To facilitate our motivation for rectifying any distorted document images,
we reformulate the pixel-wise mapping relationship between the unrestricted distorted document images and their distortion-free counterparts. Additionally, we contribute a real-world test set and new evaluation metrics to assess the quality of rectification for diverse, real-world document images. 
Through extensive experimentation on both the benchmark dataset and our own proposed dataset, our results demonstrate the efficacy and superiority of our approach.
We hope that our method can serve as a strong baseline for the community,
encouraging further research and development of practical document rectification algorithms.

{
	\bibliographystyle{IEEEtran}
	\bibliography{IEEEabrv,egbib}
}

\clearpage

\begin{IEEEbiography}[{\includegraphics[width=1in,height=1.25in,clip,keepaspectratio]{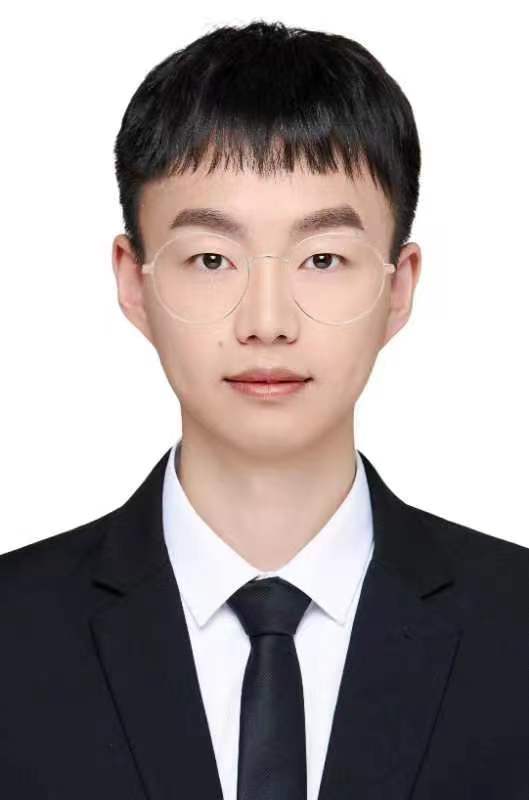}}]{Hao Feng} received the B.E. degree in electronic science and technology from Xidian University, Xi'an, China, in 2018. He is currently pursuing the Ph.D. degree in information and communication engineering with the Department of
Electronic Engineering and Information Science, USTC. His research interests include document image rectification, understanding, and computer vision.
\end{IEEEbiography}

\begin{IEEEbiography}[{\includegraphics[width=1in,height=1.25in,clip,keepaspectratio]{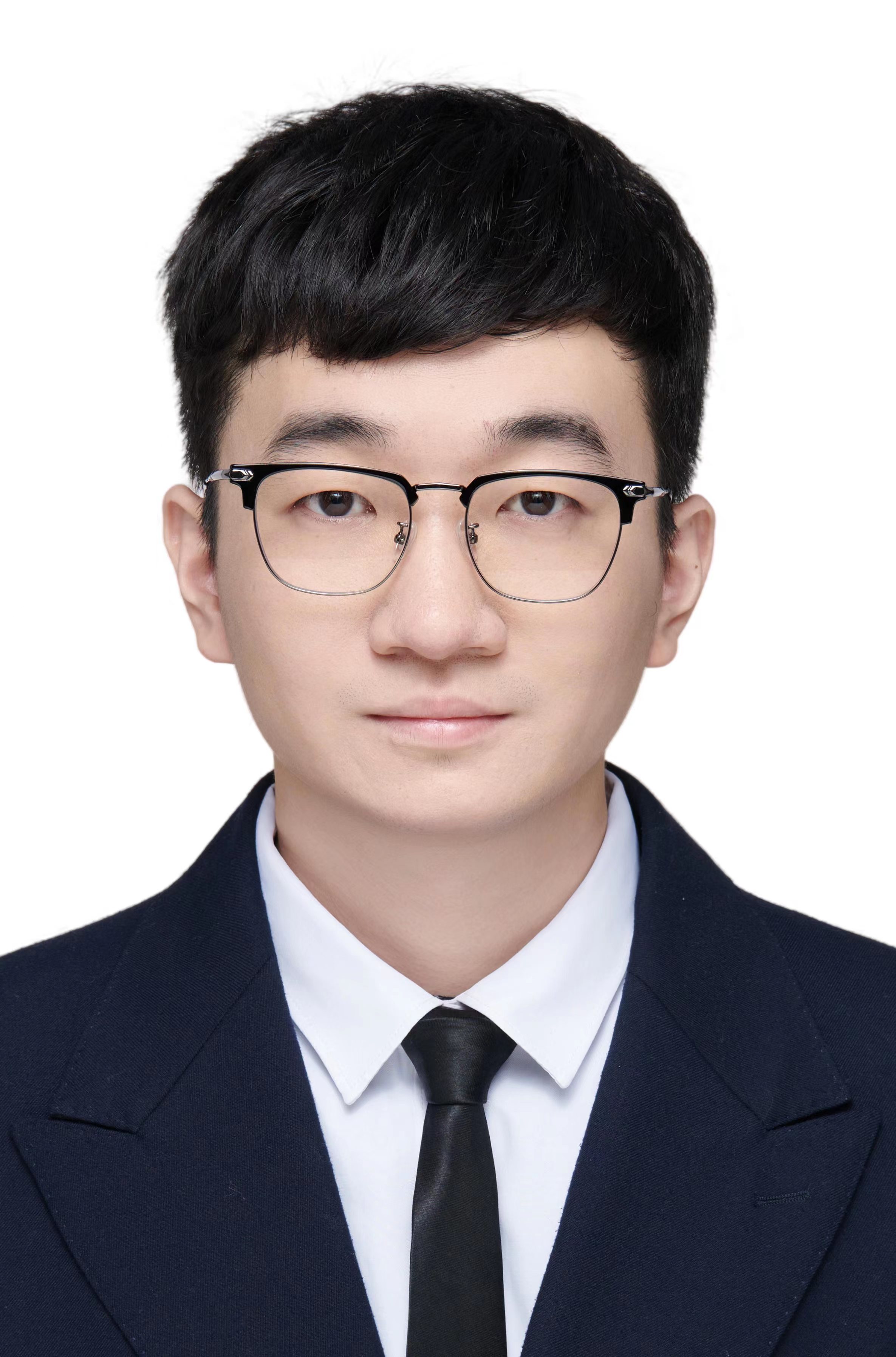}}]{Shaokai Liu} received the B.E. degree in biomedical engineering from Nanjing Medical University, Nanjing, China, in 2016. He received the M.Sc. degree in software technology from The Hong Kong Polytechnic University, Hong Kong, China, in 2018. He is currently pursuing the Ph.D. degree in the Department of Electronic Engineering and Information Science, University of Science and Technology of China (USTC). His research interests include image processing and computer vision.
\end{IEEEbiography}

\begin{IEEEbiography}[{\includegraphics[width=1in,height=1.25in,clip,keepaspectratio]{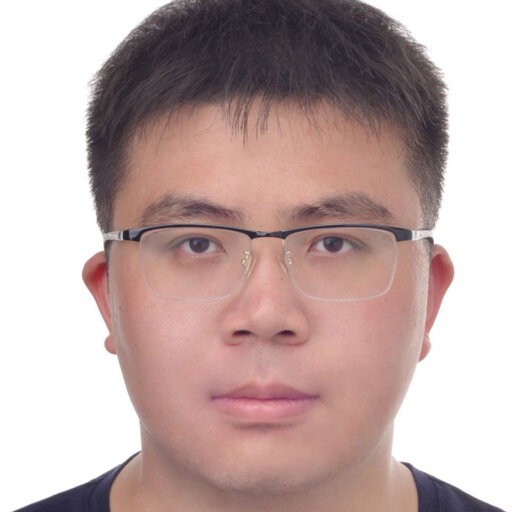}}]{Jiajun Deng} is a postdoctoral researcher at the University of Adelaide. He received his Ph.D. degree (2021) and a B.E degree (2016) from the Department of Electrical Engineering and Information Science at University of Science and Technology of China. His research interests include computer vision and multi-modal learning.
\end{IEEEbiography}

\begin{IEEEbiography}[{\includegraphics[width=1in,height=1.25in,clip,keepaspectratio]{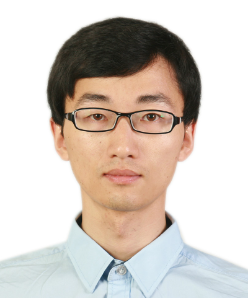}}]{Wengang Zhou} (S'20) received the B.E. degree in electronic information engineering from Wuhan University, China, in 2006, and the Ph.D. degree in electronic engineering and information science from the University of Science and Technology of China (USTC), China, in 2011. From September 2011 to September 2013, he worked as a postdoc researcher in Computer Science Department at the University of Texas at San Antonio. He is currently a Professor at the EEIS Department, USTC. His research interests include multimedia information retrieval, computer vision, and computer game. In those fields, he has published over 100 papers in IEEE/ACM Transactions and CCF Tier-A International Conferences. He is the winner of National Science Funds of China (NSFC) for Excellent Young Scientists. He is the recepient of the Best Paper Award for ICIMCS 2012. He received the award for the Excellent Ph.D Supervisor of Chinese Society of Image and Graphics (CSIG) in 2021, and the award for the Excellent Ph.D Supervisor of Chinese Academy of Sciences (CAS) in 2022. He won the First Class Wu-Wenjun Award for Progress in Artificial Intelligence Technology in 2021. He served as the publication chair of IEEE ICME 2021 and won 2021 ICME Outstanding Service Award. He is currently an Associate Editor and a Lead Guest Editor of IEEE Transactions on Multimeida.
\end{IEEEbiography}

\begin{IEEEbiography}[{\includegraphics[width=1in,height=1.25in,clip,keepaspectratio]{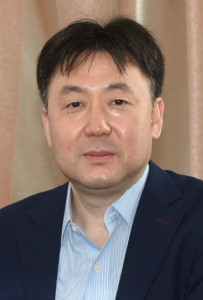}}]{Houqiang Li} (F'21) received the B.S., M.Eng., and Ph.D. degrees in electronic engineering from the University of Science and Technology of China, Hefei, China, in 1992, 1997, and 2000, respectively. He was elected as a Fellow of IEEE (2021) and he is currently a Professor with the Department of Electronic Engineering and Information Science.
His research interests include reinforcement learning, multimedia search, image/video analysis, video coding and communication, etc. He has authored and co-authored over 200 papers in journals and conferences. He is the winner of National Science Funds (NSFC) for Distinguished Young Scientists, the Distinguished Professor of Changjiang Scholars Program of China, and the Leading Scientist of Ten Thousand Talent Program of China. He is the associate editor (AE) of IEEE TMM and served as the AE of IEEE TCSVT. He served as the General Co-Chair of ICME 2021 and the TPC Co-Chair of VCIP 2010. 
He was the recipient of National Technological Invention Award of China (second class) in 2019 and the recipient of National Natural Science Award of China (second class) in 2015. He was the recipient of the Best Paper Award for VCIP 2012, the recipient of the Best Paper Award for ICIMCS 2012, and the recipient of the Best Paper Award for ACM MUM in 2011.
\end{IEEEbiography}

\end{document}